\documentclass{article}
\pdfoutput=1

\usepackage[preprint, nonatbib]{neurips_2020}

\usepackage[utf8]{inputenc} 
\usepackage[T1]{fontenc}    
\usepackage{hyperref}       
\usepackage{url}            
\usepackage{booktabs}       
\usepackage{amsfonts}       
\usepackage{nicefrac}       
\usepackage{microtype}      
\usepackage{caption}
\usepackage[table]{xcolor}

\usepackage{amsmath,bm,amsthm,bbm,amssymb,mathtools,centernot}
\usepackage{lingmacros}
\usepackage{cleveref}

\usepackage{tikz}
\usetikzlibrary{arrows}

\def\traindata{\mathcal{D}_{\text{tr}}}

\def\argmax{\ensuremath{\text{argmax}}}
\def\weights{w}
\def\x{x}

\def\likelihood{\mathcal{L}}

\newcommand{\xneg}{\x^\prime}
\newcommand{\trainneg}{\traindata^\prime}
\newcommand{\cut}[1]{}

\usepackage{graphicx,caption,subcaption}
\title{Detecting and Exorcising Statistical Demons from Language Models with
  Anti-Models of Negative Data}

\author{%
 Michael Wick \\
  Oracle Labs\\
  \texttt{michael.wick@oracle.com} \\
  \And
  Kate Silverstein \\
  Oracle Labs\\
  \texttt{kate.silverstein@oracle.com}\\
  \And
  Jean-Baptiste Tristan\thanks{Work done while at Oracle Labs.} \\
  Boston College \\
  \texttt{tristanj@bc.edu} \\
  \And
  Adam Pocock \\
  Oracle Labs\\
  \texttt{adam.pocock@oracle.com}\\
  \And
  Mark Johnson \\
  Oracle \\
  \texttt{mark.mj.johnson@oracle.com}\\
}

\begin{document}

\maketitle

\begin{abstract}
  It’s been said that “Language Models are Unsupervised Multitask
  Learners.” Indeed, self-supervised language models trained on
  ``positive'' examples of English text generalize in desirable ways
  to many natural language tasks. But if such models can stray so far
  from an initial self-supervision objective, a wayward model might generalize in
  undesirable ways too, say to nonsensical “negative” examples of {\em
    unnatural language}.  A key question in this work is: do language
  models trained on (positive) training data also generalize to
  (negative) test data?  We use this question as a contrivance to
  assess the extent to which language models learn undesirable
  properties of text, such as n-grams, that might interfere with the
  learning of more desirable properties of text, such as syntax.  We
  find that within a model family, as the number of parameters,
  training epochs, and data set size increase, so does a model's
  ability to generalize to negative n-gram data, indicating standard
  self-supervision generalizes too far.   We propose a form
  of inductive bias that attenuates such undesirable signals with
  negative data distributions automatically learned from positive
  data.  We apply the method to remove n-gram signals from LSTMs and find that
  doing so causes them to favor syntactic signals, as
  demonstrated by large error reductions (up to 46\% on the hardest
  cases) on a syntactic subject-verb agreement task.

\end{abstract}

\section{Introduction}
Some of the most elusive characteristics of language --- syntactic, semantic, and encyclopedic --- entombed in the dull workaday sentences of big text corpora, suddenly spring to life as fruitful signals when predicting masked tokens, say, the next in a language model's sequence.  And so there is a palpable optimism that self-supervising bigger models on bigger data with bigger computers will inevitably --- no, imminently --- yield systems that wield human-like language with human-like competence.  Some believe these models already have, and we just need to find evidence of it.  We're all agog as cadres of scientists eagerly seek what curious linguistic feats might be dredged up from the morass of floats and weights \cite{peters18elmo,devlin18bert,radford18improving,zhang18language,radford19language}.

But there is a problem with unrestrained self-supervision.  Namely, that big
data has bad properties too, some of which provide strong signals for language
modeling.  When Noam Chomsky was confronted with the success of big data at
a CBMM meetup at MIT \cite{chomsky17cbmm}
he warned ``big data is the wrong data'' and that
``you are not studying language, you are studying the effects of the motor
system on language,''  and that 
\begin{quote}``John and Mary {\em is} in the room'' is a more likely sentence than \\``John and Mary {\em are} in the room''\end{quote} 
thus alluding to a peculiar problem of big data with an example in want of a mull.  But let us first remind ourselves the ways in which big data is the {\em right data} by considering the sentence 
\begin{quote}Mary is in the room.\end{quote}
What might a model learn about language, from a simple sentence such as this, by merely predicting its tokens?  Suppose we try to predict the word after Mary.  Many words could possibly follow, including verbs.  How could we improve our chances of guessing the correct word?  Verbs such as ``is'' and ``travels'' can easily follow, while verbs such as ``are'' and ``travel'' cannot.   One way to exclude the second set and not the first, without resorting to brute force memorization, requires syntactic knowledge about the existence of nouns and verbs, that they have a number property which interacts with a person property, and that the subject of the sentence must agree with its verb in number.  In this way, the acquisition of syntax and semantics is a byproduct of language modeling.  Indeed, subject-verb agreement studies and others, provide evidence for this hypothesis \cite{linzen16assessing,
  kuncoro18lstms,zhang18language,warstadt19neural,wang18glue}.

Now consider how the model might predict the last word in the sentence,
``room,'' given the previous words.  Knowledge that Mary is a person, and that
people can occupy rooms and not cups, would allow the model to narrow down the
set of possible noun phrases that could end the sentence to those that could be
occupied by people.  In this way, knowledge acquisition is a byproduct of
language modeling.  The possibility of ``language models as knowledge bases''
is a captivating idea propounded in recent work \cite{petroni19language}.
We can ask the language model ``Barack Obama is married to \underline{\hspace{2em}}''
and let it fill in the answer ``Michelle''.
No
wonder ``language models are unsupervised multitask learners''
\cite{radford19language}.

Returning to mull Chomsky's example, the {\em ungrammatical} sentence has higher probability than the grammatical sentence, where we replace ``is'' with ``are.'' Why? The verb ``is'' is a more common unigram than ``are'' and ``is in'' a more common bigram than ``are in'' and so on.  Bigger data won't change these facts.  So while statistics like n-grams are often useful, they sometimes mislead.  An n-gram model is vulnerable and can be fooled up to a dependency of length {\em n}: a bi-gram can be fooled by ``Mary is'' and a tri-gram by ``Mary is in'' and so on.  But a neural language model, with its ability to capture long-distance dependencies, is vulnerable to deleterious statistical signals {\em of any length}.
Bigger models won't help. There's a sense in which the advantage neural networks have over traditional language models, their ability to model long distance dependencies, is also a weakness.  It is a weakness precisely because the data contains many undesirable signals, these signals are useful for predicting masked words, and these models have the requisite capacity and inductive bias to learn them.  But this weakness is an opportunity for improving upon the ``bigger is better'' language modeling hypothesis, if we can identify and exorcise these big data demons from our models.

It is therefore important to understand how existing models are vulnerable to
such undesireable signals, provide mechanisms for removing them, and test
if removing them improves the utility of language models by forcing them
to depend on linguistically relevant information.
We propose a method for detecting and attenuating such undesirable signals
based on negative data distributions automatically learned from positive data.
The key idea is to identify a model family that learns undesirable
characteristics of the positive data, and then use it to generate negative data
that embodies these characteristics which can be used to evaluate or train
language models.  Fortunately it's often easier to specify
models that are wrong than ones that are right. Our method turns flawed
language models, like n-grams, into signals which improve the inductive bias of
more powerful language models, like LSTMs, by guiding them away from bad
hypotheses and towards good ones.

When we apply this method to detect neural models' tendencies to learn n-gram signals, we find that as the amount of data, parameters, and training epochs increase, so does the ability to model n-gram data, indicating possible overgeneralization to the realm of unnatural language.  This is even true with GPT2 for which the larger versions of the model learn n-grams more aggressively than the smaller version.  We  then employ a loss function that allows the model to simultaneously {\em unmodel} the automatically generated negative data while modeling the observed positive data.  We find the method successfully attenuates n-gram signals and that doing so improves the models' ability to capture syntactic properties, achieving a 46\% error reduction in a subject-verb agreement task \cite{linzen16assessing}.

\section{Anti-models for detecting and attenuating properties of text}
Our approach
for detecting and attenuating undesirable signals is a
process we term {\em anti-modeling}.  The idea is to select a
model that has a particular flaw (e.g., due to an incorrect assumption about
the data, like a Markov assumption) such that the flaw makes
the model vulnerable to some inherent undesirable characteristic of the
data.  Then, fit the flawed model to the data, to create a ``negative data
distribution'' that we use to generate negative data for
detecting and attenuating signals.  A diagram sketching how the procedure is
used for training with attenuation is in Figure~\ref{fig:app-exor-diagram} in the appendix.

In this work, we focus on n-grams, which are extremely useful,
but ultimately imperfect accounts of language \cite{chomsky57syntactic}.   The Markov assumption in an n-gram model means it has a limited view of language but once trained we can generate negative data (e.g.,\ from a tri-gram model of Penn Tree Bank):
\begin{quote}
the organization 's desire for pork tends to lock in business
long-term but cut revenue in transactions for its premium products
\end{quote}
While somewhat nonsensical, there are some real senses in which this negative
sentence is reasonably language-like.  Box's aphorism that ``all models are
wrong, some are useful'' certainly applies to the very n-grams we propose to
remove.  Indeed, studies on locality effects in human sentence processing show
that close by dependencies are easier to process
\cite{gibson1998linguistic,demberg2008data,Futrell2017}; to maximize
communicative efficiency, humans are predisposed toward producing utterances 
that are easy for an n-gram language model to fit. Of course, n-gram
language models are not perfect models of natural language, because in
practice humans use both external context and longer range
syntactic structures to communicate information.  When applying our
method to attenuating n-gram statistics, it serves as a regularizer to
help the model avoid local minima that occur due to overfitting n-grams, allowing
it to learn long-distance dependencies. There is a
danger that training the model to ignore the n-gram statistics will
cause it to poorly fit an otherwise straightforward sentence, we investigate
this in Section~\ref{sec:experiments}.

\subsection{Evaluating good models with bad ones}
In order to employ anti-models for evaluation, we first fit the model
to either the test or development data and then employ the model to
automatically generate negative versions of those datasets.  Once we
have a negative data distribution, we can evaluate a language model's
ability to model that data by measuring perplexity.  Our assumption is
that the better the model fits the negative evaluation data, the more
strongly the model has internalized the undesirable property that the
negative data embodies.  In the evaluation, a good language model
should ascribe high probability (low perplexity) to the positive data
and ascribe low probability (high perplexity) to the negative data.
The gap between the positive and negative perplexities can summarize
the evaluation.  We use our method as a tool, in
Section~\ref{sec:experiments}, to study the extent to which various
language models learn n-grams from positive training
data.

\subsection{Training good models from bad ones}
Most generative machine learning algorithms work only with ``positive'' data
for which the goal is to find the hypothesis class that has the highest data
likelihood on said data, possibly with some regularization.  Here we describe a
method for incorporating {\em negative data} of the form previously discussed.
The idea is to jointly maximize the likelihood of the positive data, 
and minimize the likelihood of the negative data.
In this way, negative data coming
from our flawed models of language can guide the more powerful models away from
these flaws and towards better hypotheses.  We term this procedure anti-modeling.

\cut{
\paragraph{Anti-models of negative data}
First, we must determine an appropriate anti-model or negative data
distribution from which we can (a) generate negative data that imbues
the appropriate inductive bias when unlearned by our actual model and
(b) learn the model from the available positive data.  More formally,
let $\mathcal{M}$ be the original model we wish to train.  

\paragraph{Unlearning negative data}
}

But how should we accomplish this?  It turns out that this is not completely
straightforward.  One challenge is that if we subtract the negative from the
positive likelihood their arithmetic interaction blocks the log from
telescoping into the negative likelihood products, rendering stochastic
gradient descent (SGD) impossible at a per example level.  On the other hand if
we perform the subtraction in log space --- and are not careful --- it results
in a degenerate loss function that is unbounded and cannot be optimized.  We
develop these points more in this section and propose a particular objective
function that (a) incorporates negative data and (b) allows for per-example SGD
in log space.

Formally, if we train a neural LMs weights $\weights$ by maximizing the
likelihood $\mathcal{L}$ of the training data $\traindata$
\begin{equation}
\hat{\weights} = \underset{\weights}{\argmax}\likelihood(\traindata,\weights)
\end{equation}
then a reasonable objective function for including negative data $\trainneg$ is 
\begin{equation}
  \label{eqn:obj-neg0}
\hat{\weights} =
\underset{\weights}{\argmax}\;\likelihood(\traindata,\weights)\left(1 - \likelihood(\trainneg,\weights)\right)^\alpha
\end{equation}
where the $\alpha$ hyperparameter governs the relative strength of the negative data.

For gradient-taking expedience, it is often desirable to work in log
space but note that Equation~\ref{eqn:obj-neg0} is mathematically
inconvenient, for if we attempt to take the log, it is abruptly
blocked by the one-minus-likelihood term, making it difficult to apply
SGD on a per-example level.  However, if we first expand out the
definitions of likelihood as a product over all sequences $\x$ in the
training data $\traindata$
\begin{equation}
\hat{\weights} =
\underset{\weights}{\argmax}\left(\prod_{\x\in\traindata}\prod_{\x_i\in
    x}P(\x_i|\weights)\right)  \left(1-  \prod_{\xneg\in\trainneg}\prod_{\xneg_i\in\xneg}P(\xneg_i|\weights) \right)^\alpha
\end{equation}
and push the one-minus term inside the product, then we can lower bound it via Jensen's inequality
\begin{equation}
  \label{eqn:obj-neg2}
\hat{\weights} =
\underset{\weights}{\argmax}\left(\prod_{\x\in\traindata}\prod_{\x_i\in
    x}P(\x_i|\weights)\right)
\left(\prod_{\xneg\in\trainneg}\prod_{\xneg_i\in\xneg}1 - P(\xneg_i|\weights) \right)^\alpha
\end{equation}
So rather than maximizing ``one-minus-the-likelihood'' for the
negative data, we maximize the product of
``one-minus-the-probability'' of each token in the
dataset.  Now
unfettered by addition, the log loss for a single
negative sentence $\xneg$ is readily available
\begin{equation}
  -\alpha \sum_{\xneg_i\in\xneg}\log(1 - \exp(-\ell(\weights,\xneg_i)))
\end{equation}
where $\ell(\weights, \xneg_i)$ is the usual cross entropy
loss term, but taken between the model with weights $\weights$ and a
token in the negative data
sequence $\xneg_i\in \xneg \in \trainneg$.
This loss has nearly the same form as the ``unlikelihood,''  a method for
reducing redundancy of neural text generators
\cite{welleck2019neural}.

\section{Related Work}

Negative data has a storied history in linguistics, natural language processing
(NLP), and machine learning. In linguistics, it sometimes appears in the
innateness debate since negative data could be a solution to phenomena like
Baker's paradox, but the role and quantity of negative data in a child's
language learning is questionable
\cite{yang16price,baker1979syntactic,lawrence2000,elman1998rethinking}.  Gold's
theorems demonstrate, under certain assumptions, that learning a language is
impossible from positive data alone~\cite{gold1967language}.  Methods that take
into account negative data to learn artificial languages are usually
discriminative and classify a sentence as to whether or not it is part of the
language \cite{lawrence2000}.  In contrast, we directly unmodel the negative
data in a more generative way. Contemporaneously, negative data is being
proposed for models of artificial language but could not apply to natural
languages because it relies upon knowledge of the true grammar to
generate negative data \cite{noji2020analysis}.  In contrast, our method is an
automated way of generating negative data for natural languages for which
we do not know the grammar.

Recent work in text generation employs a loss function called the
``unlikelihood'' that has the same form as the particular loss we present in
this work \cite{welleck2019neural}.  The goal in that work is to reduce the
repetitiveness and increase the quality of generated output and the
unlikelihood penalizes such generated outputs at train time.  In contrast, our
use of the loss is for unmodeling our automatically generated negative data as
a form of inductive bias for language models.  Other work, that employs
discriminators to enforce Grice's maxims during language generation, or work
that enforces hard constraints on the output of sequence-to-sequence models is
also relevant~\cite{holtzman2018learning, lee19gradient}.

Our work is closely related to model distillation and data augmentation~\cite{hinton2015distilling}.  In distillation a small student model is taught to mimic a large teacher model using some training dataset.  Our training method is essentially an anti-distillation process: the n-gram model is an {\em anti}-teacher or bad role model that the student actively avoids.  In data augmentation, additional (positive) data is created by transforming the input, with the aim to express an invariant (\emph{e.g.}\ to language in multilingual word embeddings \cite{wick16minimally}).  Recently explicit syntactic data augmentation has been applied to improve syntax learning in masked language models \cite{min2020syntactic}.  We can see our approach as a twisted form of data augmentation with a special loss function that unlearns the augmented data, rather than learning it.

It is tempting to make a connection between our training technique and generative adversarial networks (GAN), which are not well suited for text data.  However, there are several fundamental differences that give our method a greater chance for success.  First, we make no attempt to learn an adversarial generator.  Instead, we employ a negative generator that is fixed beforehand and acts as a guide of what not to do. This allows us to control both what the language model learns (via training data) and what it does not learn (via the generator's design).  Moreover training on generated output for LMs is generally hard~\cite{bengio15scheduled,chang15learning}.  Second, the dual generator design means we no longer have to rely on the meager bit of information that a binary discriminator imparts upon the weights in each update.  Instead, every token in the sequence imparts $\log(V)$ bits (where $V$ is the vocabulary size).

Studying the capability of neural language models is important: syntactic ability, acceptability judgements, and usefulness for downstream tasks \cite{linzen16assessing,wang18glue,warstadt19neural}.   These evaluations are key for understanding the desirable ways in which such models generalize, while our anti-modeling evaluation is complementary and helpful for understanding how such models might generalize too far.

\section{Experiments}
\label{sec:experiments}
In our experiments we evaluate and train language models with the
anti-modeling approach and see how they might improve.  First, we
employ the anti-models as an evaluation tool to study the extent to
which language models learn n-gram statistics as we vary
the number of parameters, training epochs, data size, and model family
(recurrent vs transformer).  Second, we employ the anti-models to
generate negative training data for attenuating n-gram statistics from
language models, and we
investigate how it affects the perplexity on both the positive and negative data.
Third, we see if attenuating n-gram signals allows the model to learn
more syntax with subject-verb agreement tasks.

\subsection{Data and models}
\label{sec:data}
\paragraph{Data}
We employ a pre-processed
version of the Penn Tree Bank (PTB) \cite{marcus1993building} and the subject-verb agreement
data \cite{linzen16assessing}.  Both have a vocabulary of around 10K
words, with rare words mapped to an unknown token (in PTB) or their
part of speech (in subject-verb).  Details are in
Appendix~\ref{app:data}.

\paragraph{Negative data}
We focus on word-based tri-grams as our negative data model
because we reasoned that they are a sweet spot for our dataset size:
it is the largest model we thought would not overfit the
corpus and generate `negative' sentences that were identical or too
similar to the positive training sentences. In Appendix~\ref{app:n-gram} we
present experiments with bi-grams and 4-grams.
We fit the tri-grams to the various positive datasets with unsmoothed
maximum likelihood.  At train time we generate data on the fly for
each positive example, using the previous sentence as context for
generation.  For validation, we generate static datasets with the
tri-gram model trained on positive development data.

\paragraph{Models}
We primarily employ word-level recurrent networks (LSTMs following
\cite{zaremba14recurrent}) but we also test the transformer-based GPT2
\cite{radford19language} in some of our experiments. For the former, we
replicate their three LSTMs of three different sizes (200 hidden units, 650
hidden units, and 1500 hidden units). For the latter, we
evaluate pre-trained GPT2 models of different sizes (117M, 345M, 774M and 1558M
parameters) using our method. We explore fine-tuning GPT2 with negative data in
additional experiments presented in Appendix~\ref{app:gpt2}. For the LSTMs, we
employ negative log likelihood training as a baseline and refer to this as
``nll.''  When we employ negative data, with a given weight $\alpha$ on
negative portion of the loss, we refer to this as ``nll-negative-$\alpha$''
with some value of $\alpha$.

\subsection{Results: detecting and attenuating n-grams}
\label{sec:detecting}
\begin{figure}[t]
  \centering
\begin{subfigure}[b]{0.32\textwidth}
  \centering
  \includegraphics[width=\textwidth]{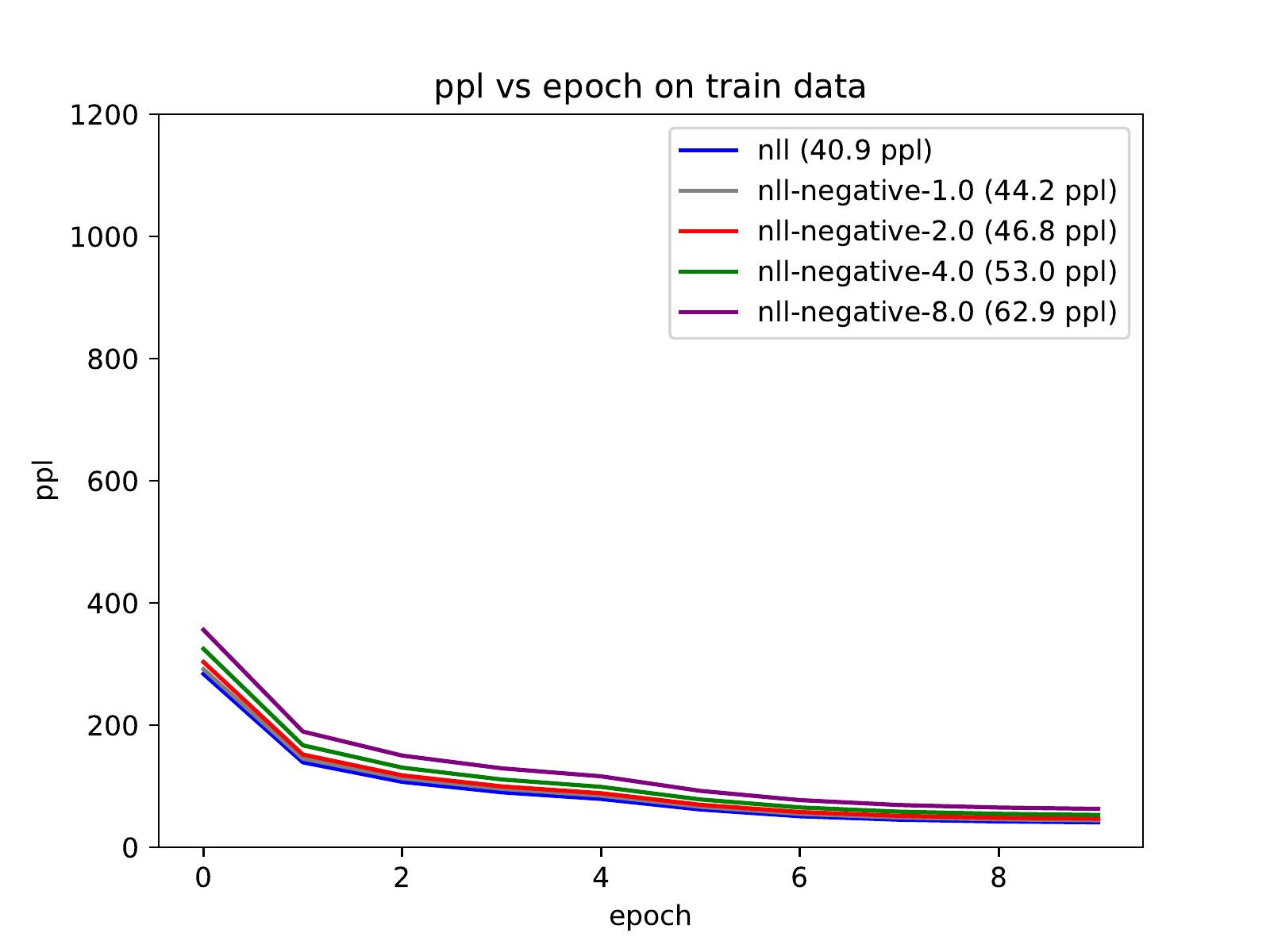}
  \caption{Train perplexity.}
  \label{fig:train-ppl}
\end{subfigure}
\begin{subfigure}[b]{0.32\textwidth}
  \centering
  \includegraphics[width=\textwidth]{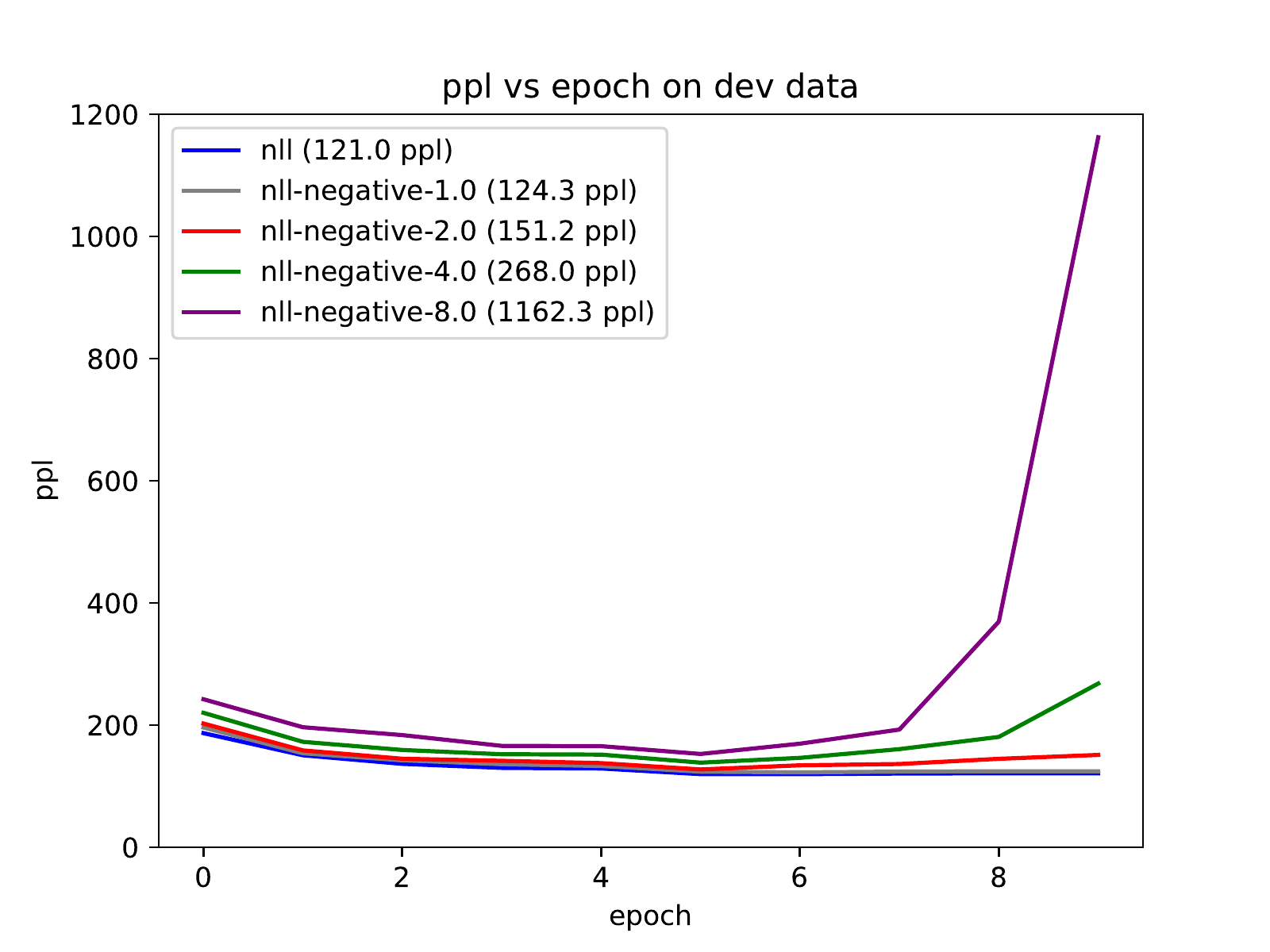}
  \caption{Dev perplexity (positive data).}
  \label{fig:dev-ppl}
\end{subfigure}
\begin{subfigure}[b]{0.32\textwidth}
  \centering
  \includegraphics[width=\textwidth]{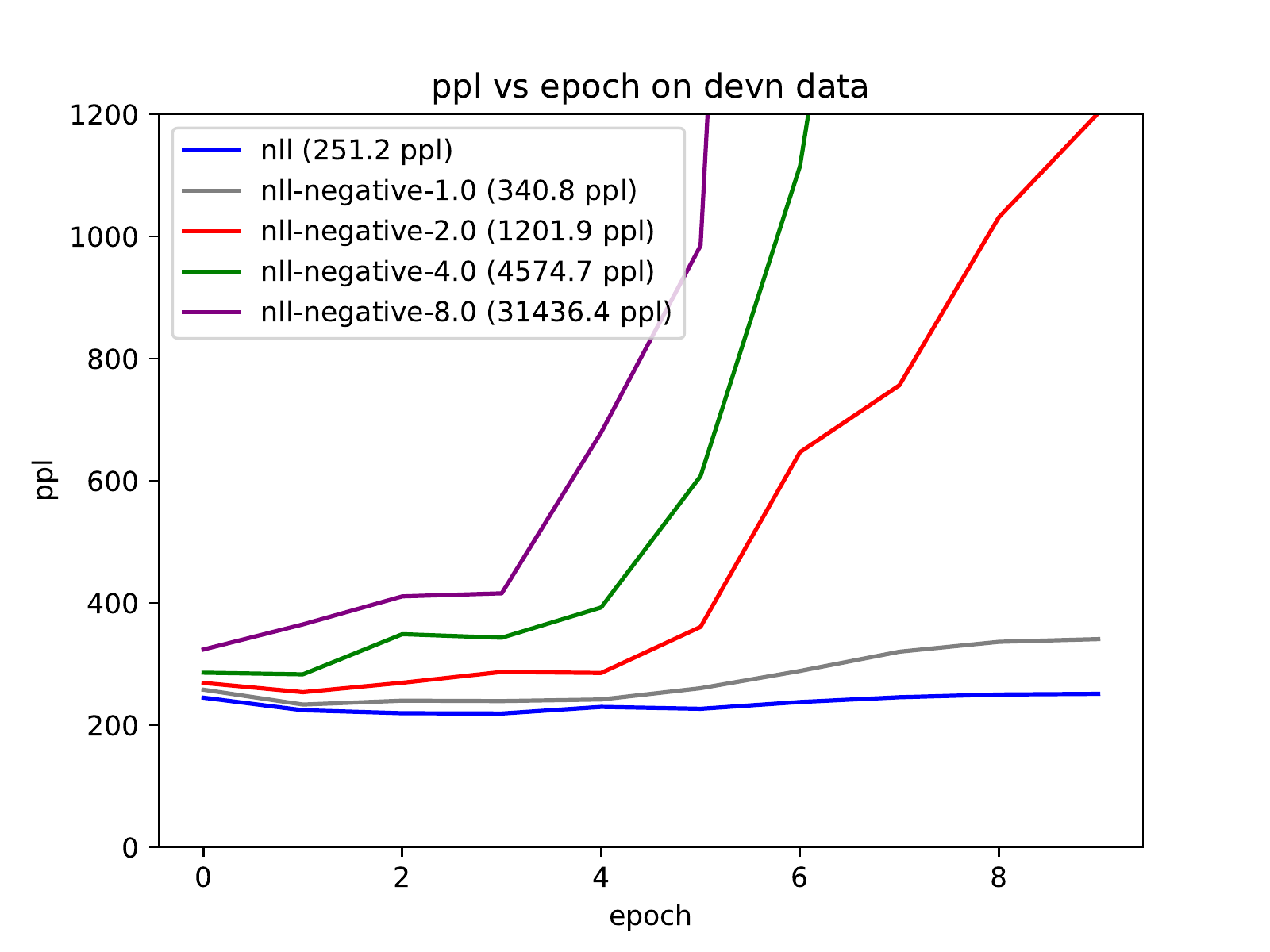}
  \caption{Dev perplexity (negative data).}
  \label{fig:devn-ppl}
\end{subfigure}
\caption{Perplexity on train, dev, and negative dev data for
various training methods (small LSTM).
\label{fig:lstm-small-ppl}}
\centering
\begin{subfigure}[b]{0.32\textwidth}
  \centering
  \includegraphics[width=\textwidth]{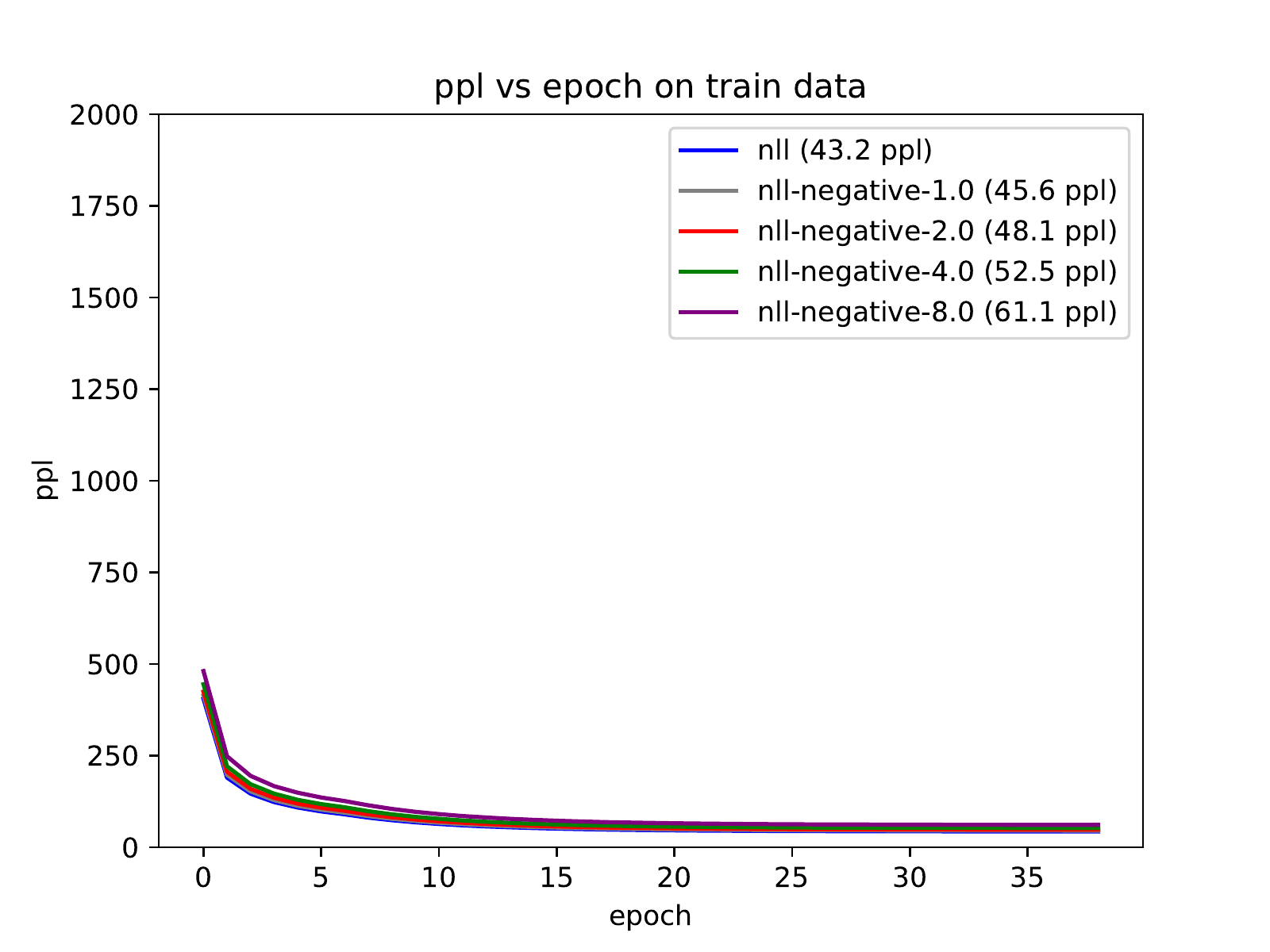}
  \caption{Train perplexity.}
  \label{fig:lstm-medium-train-ppl}
\end{subfigure}
\begin{subfigure}[b]{0.32\textwidth}
  \centering
  \includegraphics[width=\textwidth]{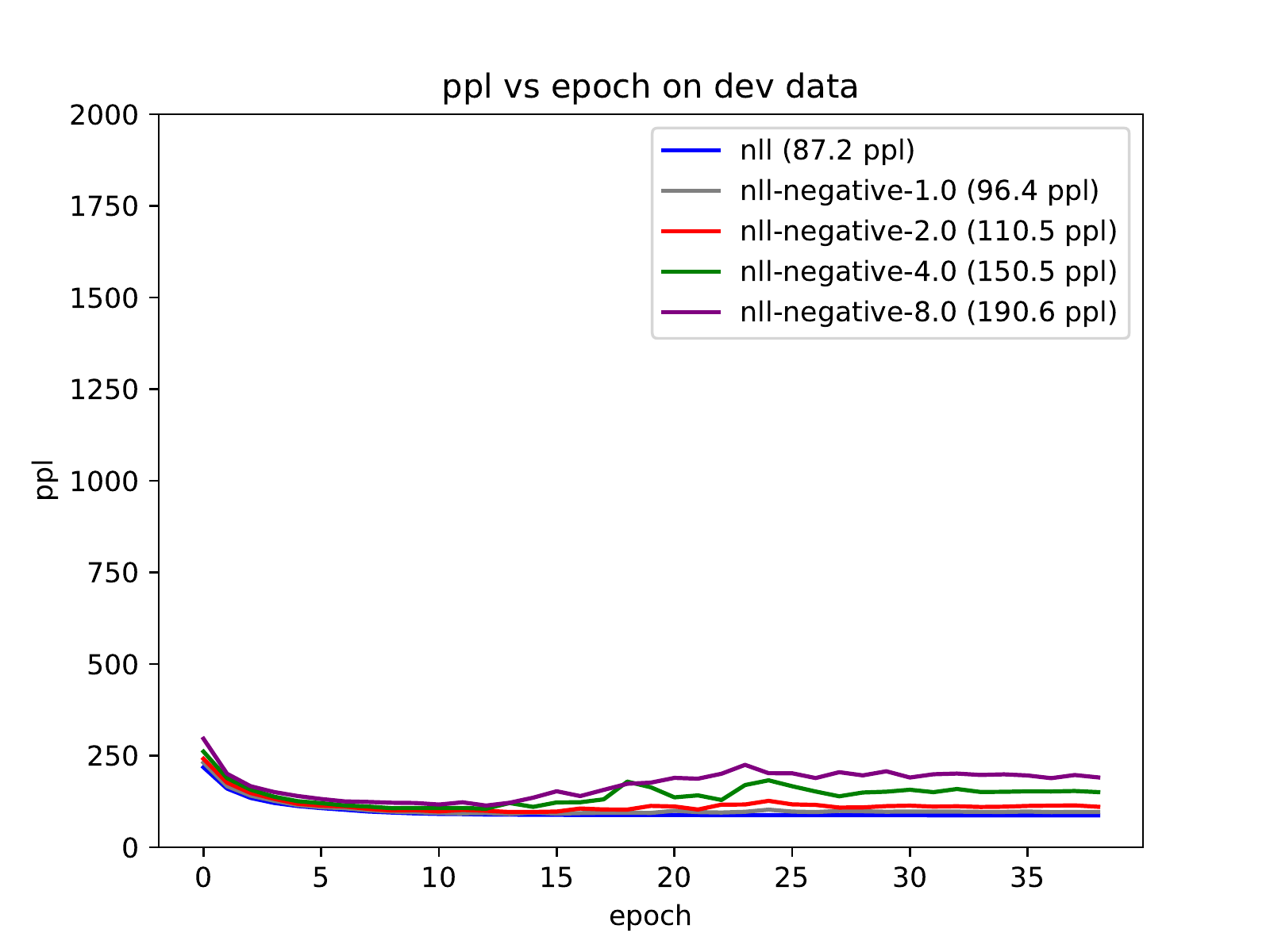}
  \caption{Dev perplexity (positive data).}
  \label{fig:lstm-medium-dev-ppl}
\end{subfigure}
\begin{subfigure}[b]{0.32\textwidth}
  \centering
  \includegraphics[width=\textwidth]{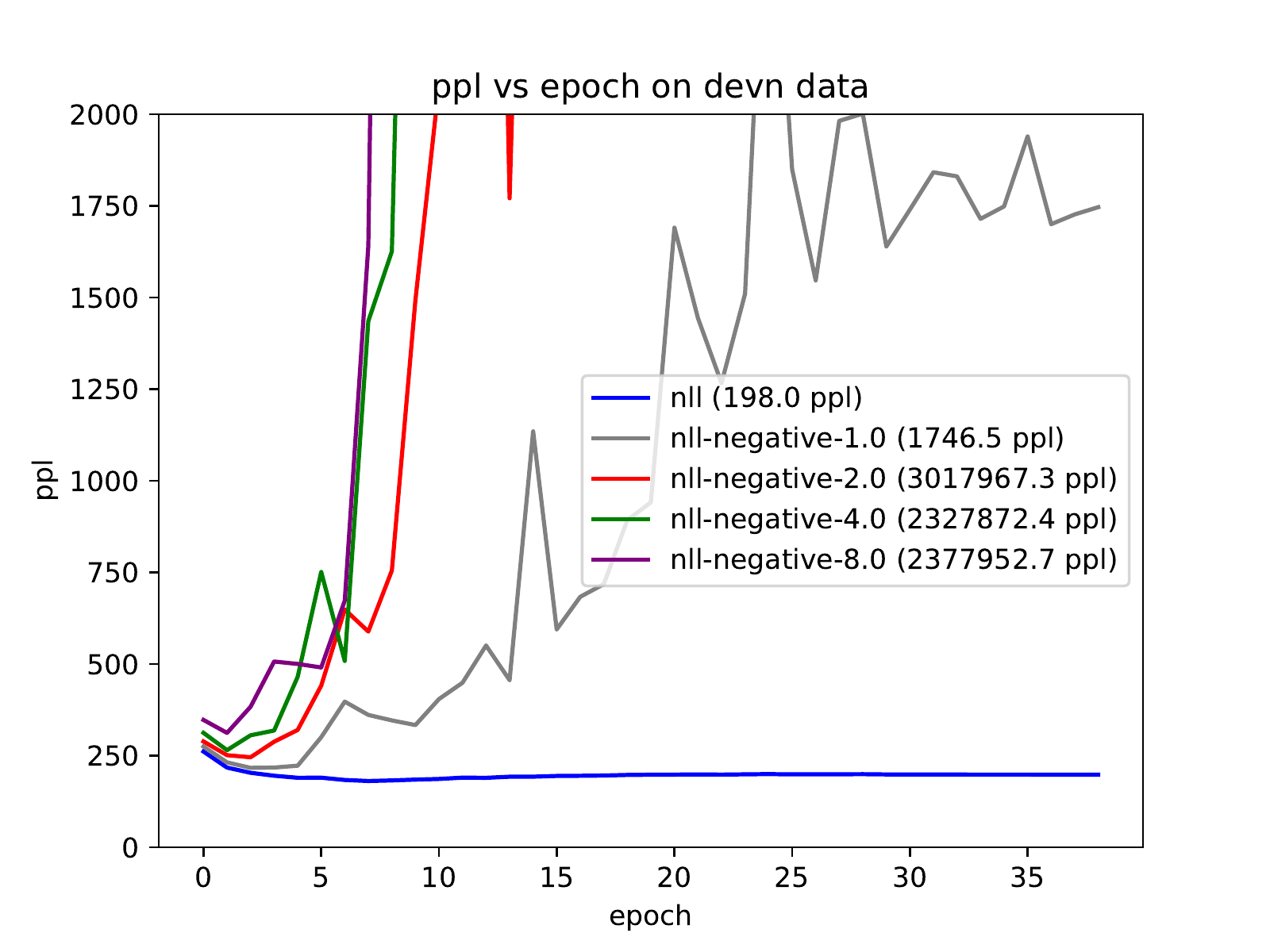}
  \caption{Dev perplexity (negative data).}
  \label{fig:lstm-medium-devn-ppl}
\end{subfigure}
\caption{Perplexity on train, dev, and negative dev data for
various training methods (med LSTM).
\label{fig:lstm-medium-ppl}}
\centering
\begin{subfigure}[b]{0.32\textwidth}
  \centering
  \includegraphics[width=\textwidth]{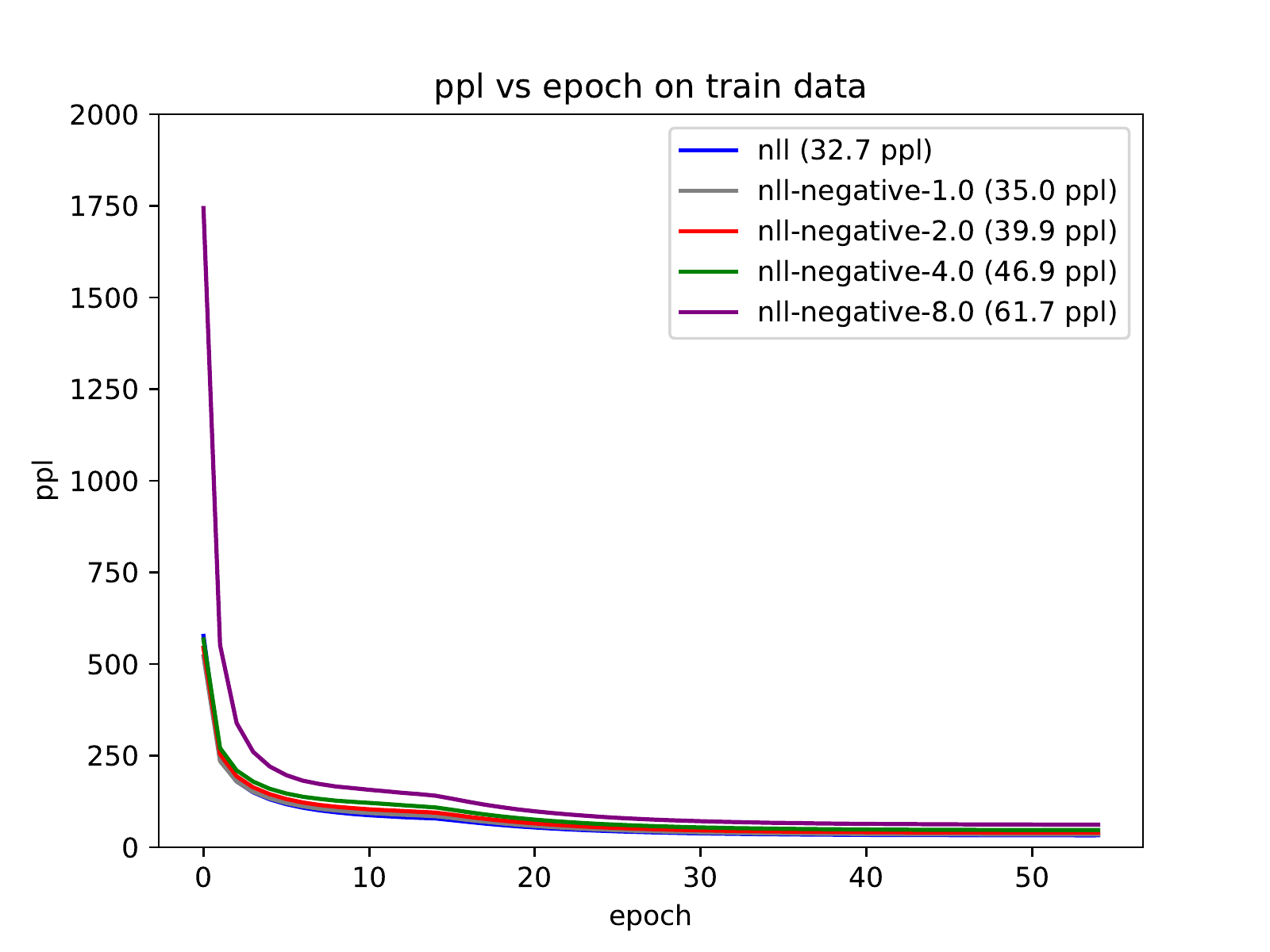}
  \caption{Train perplexity.}
  \label{fig:lstm-large-train-ppl}
\end{subfigure}
\begin{subfigure}[b]{0.32\textwidth}
  \centering
  \includegraphics[width=\textwidth]{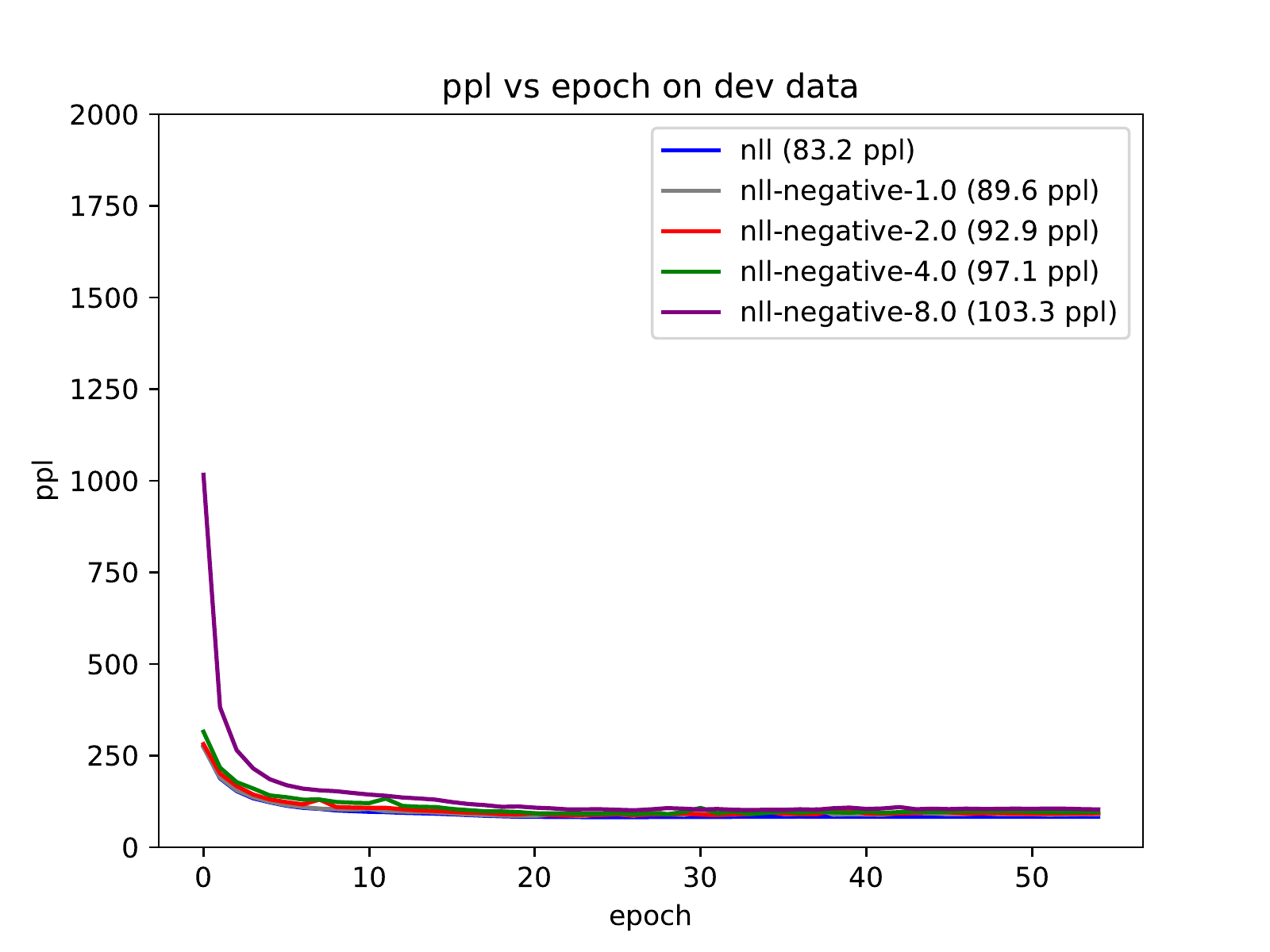}
  \caption{Dev perplexity (positive data).}
  \label{fig:lstm-large-dev-ppl}
\end{subfigure}
\begin{subfigure}[b]{0.32\textwidth}
  \centering
  \includegraphics[width=\textwidth]{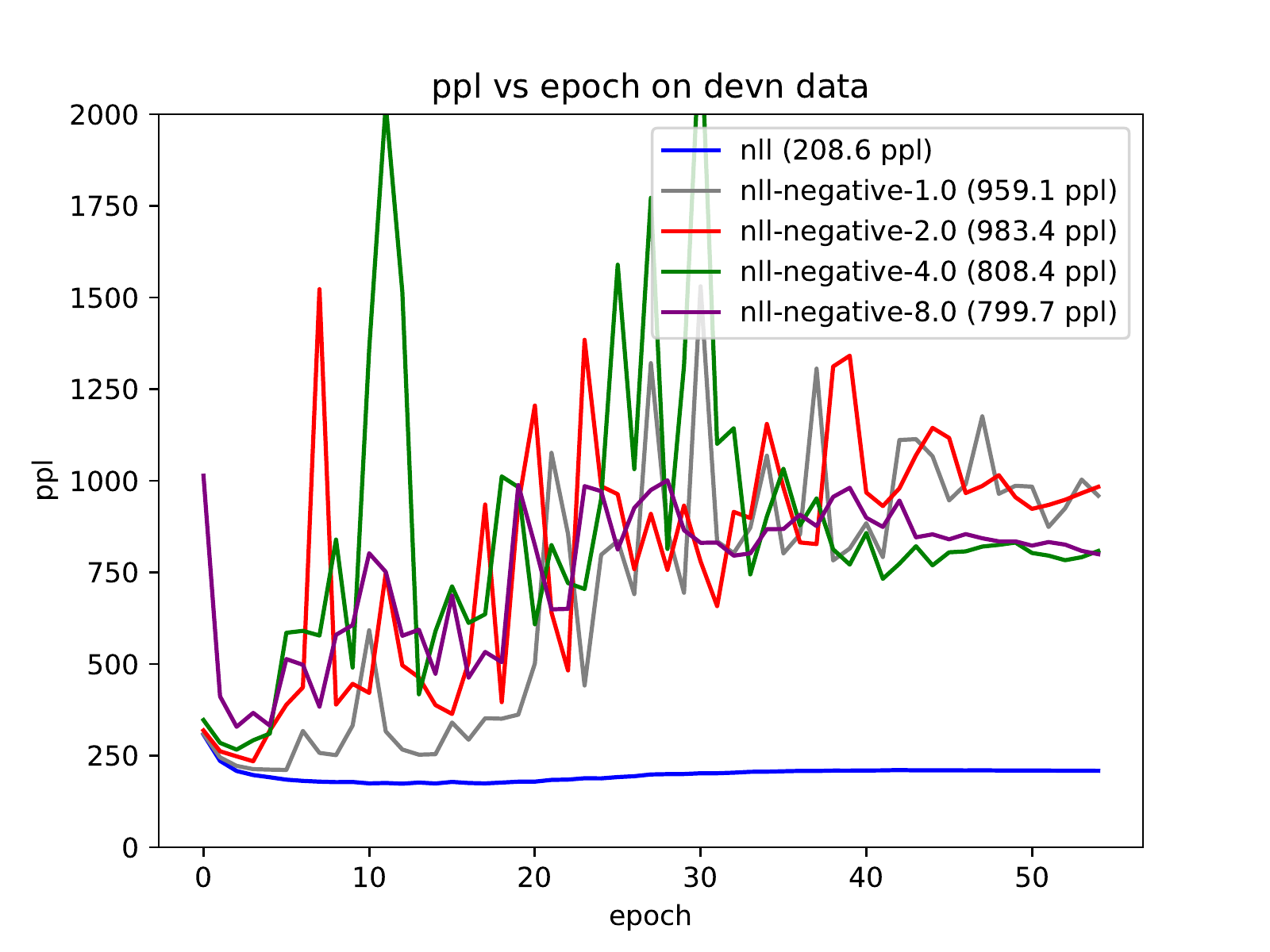}
  \caption{Dev perplexity (negative data).}
  \label{fig:lstm-large-devn-ppl}
\end{subfigure}
\caption{Perplexity on train, dev, and negative dev data for
  various training methods (large LSTM).
  \label{fig:lstm-large-ppl}}
\end{figure}

\begin{figure}
\begin{subfigure}[b]{0.32\textwidth}
  \centering
  \includegraphics[width=\textwidth]{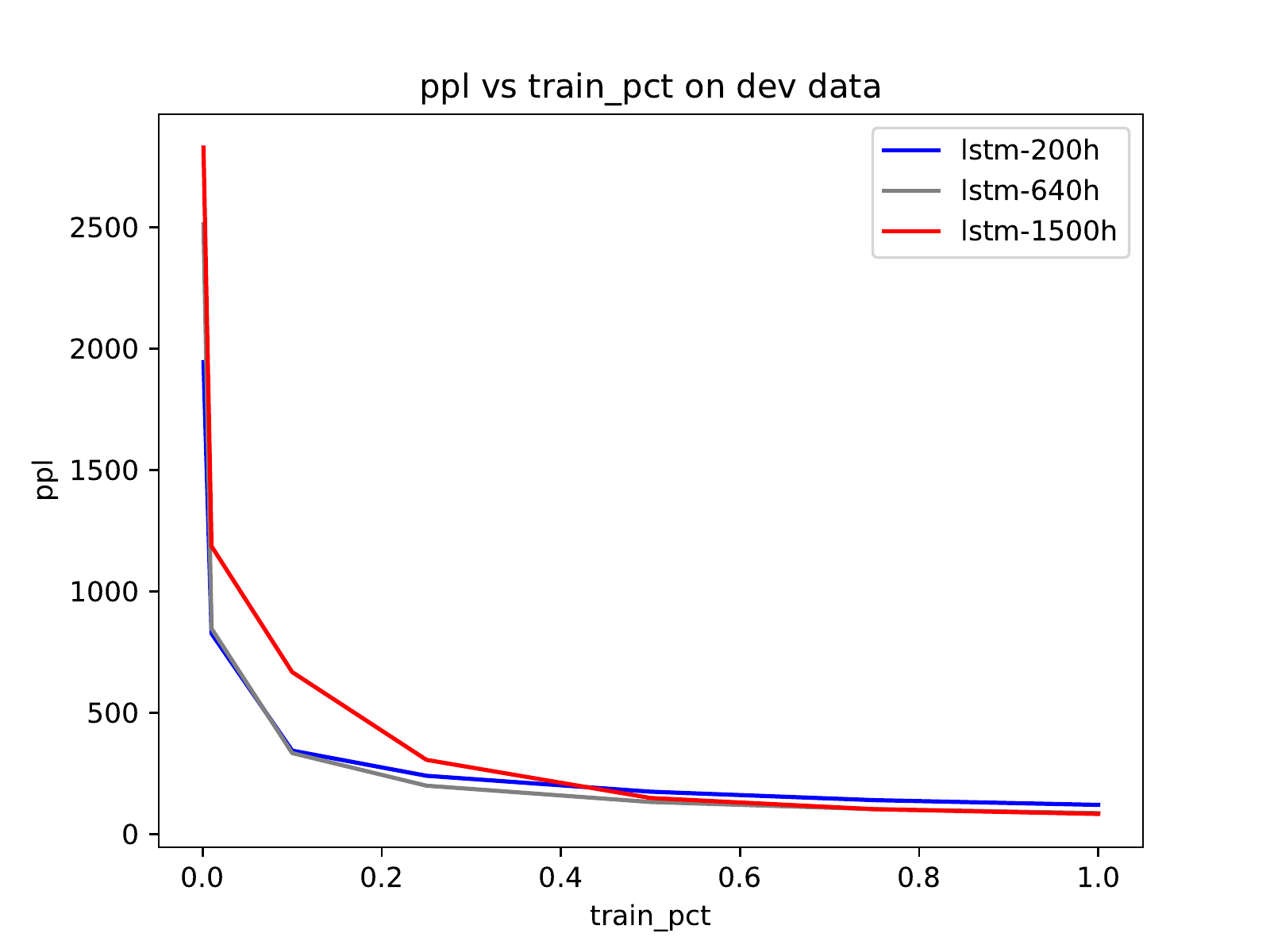}
  \caption{Positive perplexity.}
  \label{fig:data-curve-positive}
\end{subfigure}
\begin{subfigure}[b]{0.32\textwidth}
  \centering
  \includegraphics[width=\textwidth]{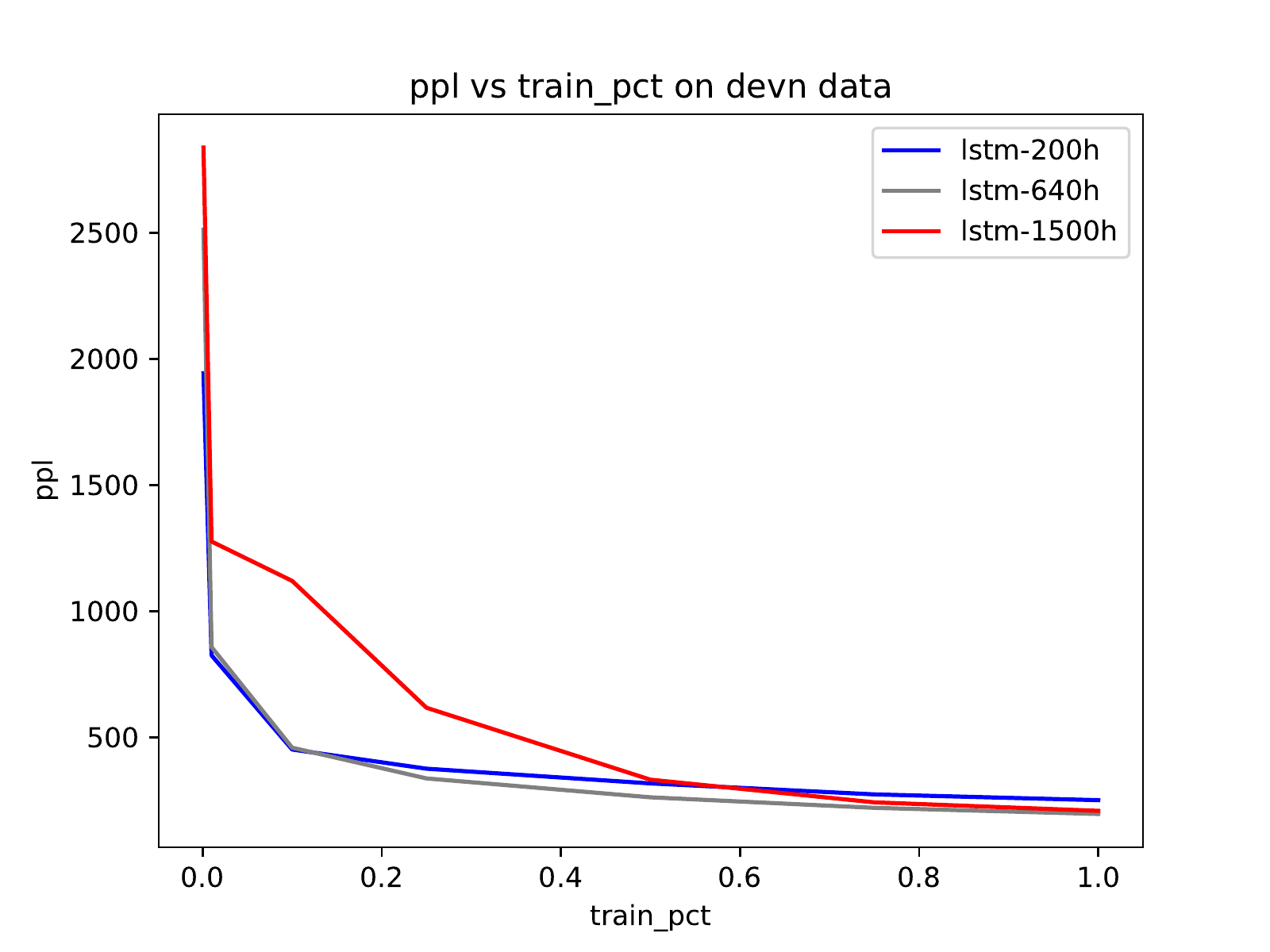}
  \caption{Negative perplexity.}
  \label{fig:data-curve-negative}
\end{subfigure}
\begin{subfigure}[b]{0.32\textwidth}
  \centering
     \includegraphics[width=\textwidth]{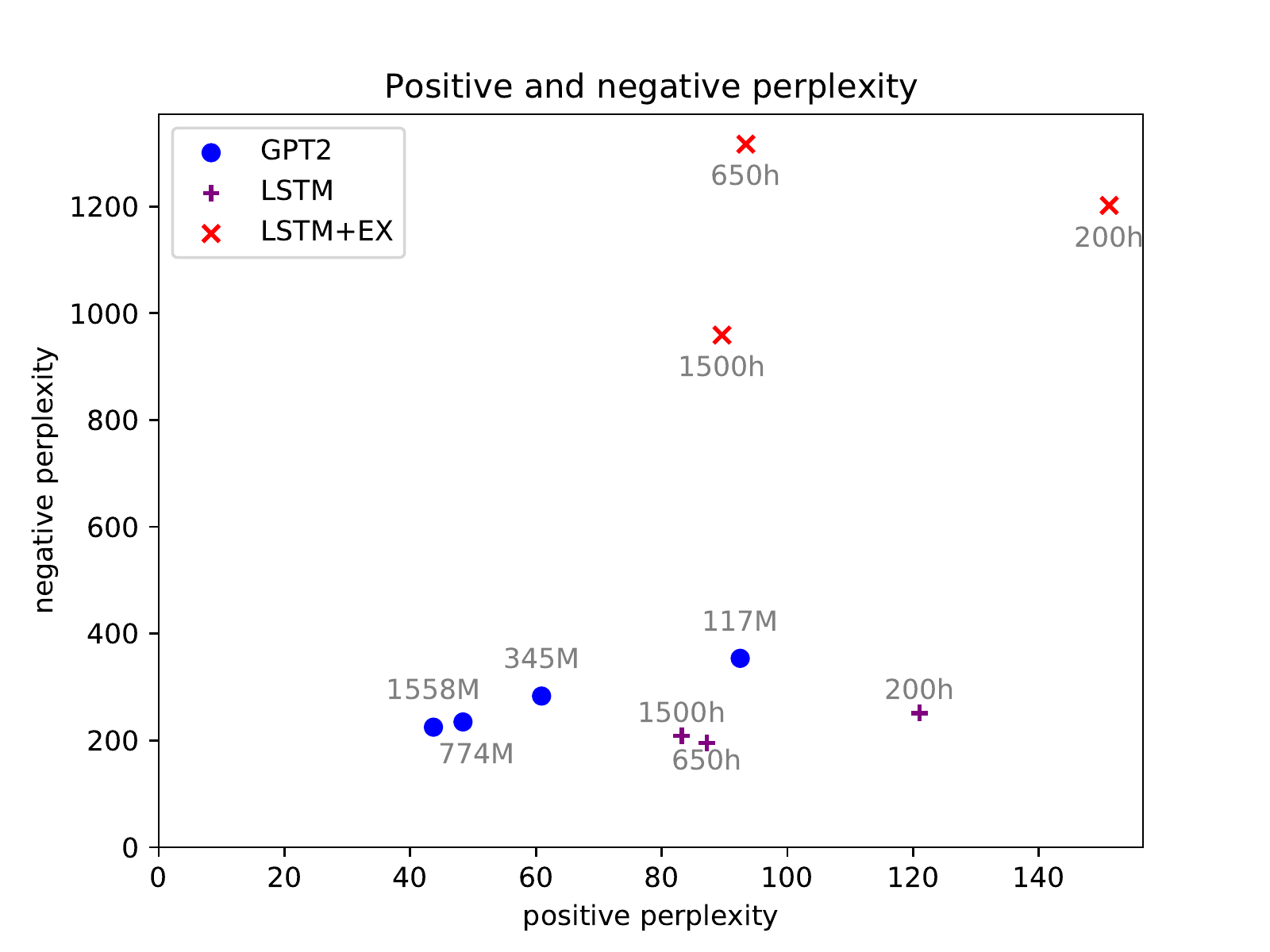}
\caption{Positive vs negative perplexity.}
   \label{fig:neg-vs-pos}
\end{subfigure}
\caption{(a) \& (b): the positive and negative effects of more
  training data.   Portion of training data increases along x-axes,
  and y-axes is perplexity. (c): positive vs negative ppl for
  various models.}
  \label{fig:data-curve}
\end{figure}

Here we study the extent to which a model learns undesirable tri-gram
signals with negative data.  We evaluate how well various language
models perform on both positive and negative data.  A language model
should not only faithfully model the positive data, but it should also
fail to model the negative data.  That is, the best language models
should ascribe higher probability (lower perplexity) to positive data
and ascribe lower probability (higher perplexity) to negative data. We
train each LSTM model (small, medium, and large) while varying
$\alpha\in\{0, 1, 2, 4, 8\}$ for each of the three LSTMs ($\alpha=0$
is the traditional LSTM model in which we train with negative
log-likelihood on the positive data).  We plot the results in three
figures, corresponding to small (Figure~\ref{fig:lstm-small-ppl}),
medium (Figure~\ref{fig:lstm-medium-ppl}), and large
(Figure~\ref{fig:lstm-large-ppl}) LSTMs.  Each figure contains three
plots, demonstrating the models' perplexity results, after each epoch,
on the positive training data, the positive dev data, and the {\em
  negative} dev data.  We also report the final perplexity that each
model achieves in the plot legends.

First, focusing on the ``nll'' baselines, which are LSTMs trained only on the
positive data with negative log likelihood, we see that for all three models
the perplexity on the positive development data improves over time, as
expected.  However, the negative perplexity also improves: additional training
epochs allow the three models to (undesirably) learn the negative tri-gram
data.  Fortunately, this behavior does not last forever, and the models begin
to recover slightly (after about 15 epochs for the large LSTM). Looking across
the plots for the three LSTM sizes, as the number of parameters increases (to
200, 650 and 1500 hidden units), we see that positive perplexity decreases, as
expected.  However, the increase in capacity also allows the model to fit the
negative data better, providing evidence for an increased reliance on n-gram
signals.  Finally, in
Figures~\ref{fig:data-curve-positive}\&\ref{fig:data-curve-negative}, as we
vary the amount of positive training data, we observe a decrease
in both positive and negative perplexity.

Our results indicate for models trained {\em only on positive
  data}, that increasing the amount of computation, the amount of model
capacity, and the amount of positive training data, increases the
model's ability to fit both the positive {\em and negative data.}  Of
course, one explanation might be that n-grams are highly correlated
with real English text, and there is a case to be made for this.
However, examining sample sentences (some shown in the appendix in
Figures~\ref{fig:negative-examples-dev-ptb}\&\ref{fig:negative-examples-dev-sva}),
indicates that these sentences are different enough from natural
language to quell major concerns.  More importantly, if n-grams and
English text were too tightly correlated, then it should be difficult
to unmodel the n-grams without unmodeling the positive English text.
So what happens when we employ negative data at train time?

Revisiting
\Cref{fig:lstm-small-ppl,,fig:lstm-medium-ppl,,fig:lstm-large-ppl}, 
we see for all three LSTMs increasing the weight on the negative term
of the loss function causes the models to aggressively
unmodel the negative data.  While the perplexity also increases on the
positive data, it does not increase nearly as dramatically as on the
negative data, except for the small LSTM when using a large 
weight on the negative term.  
Hence we can successfully remove the n-gram
signal without damaging the overall language model.

We also plot the positive (x-axis) and negative (y-axis) perplexities in a
scatter plot for each of the above models, as well as four transformer models
(GPT2) of different sizes\cite{radford19language}.  The intent is to see, for
each model family, how changes in positive perplexity correspond to changes in
negative perplexity.  What we observe (Figure~\ref{fig:neg-vs-pos}) is that for
each model family, when trained only on positive data, that as the positive
perplexity decreases, so does the negative perplexity.  While the transformers
have a slightly higher perplexity on negative data than the LSTMs --- perhaps
not surprising given the known recency bias of LSTMs \cite{ravfogel19studying}
--- they also exhibit this trend.  However note that when we train the LSTMs
with the negative data, they have far worse perplexity on the negative data
than even the transformers, while their perplexities on the positive data is
only slightly decreased.

\subsection{Results: attenuating n-grams improves syntax in language
  models}
\label{sec:sva}
\begin{figure}
    \centering
  \begin{subfigure}[b]{0.48\textwidth}
   \includegraphics[width=\textwidth]{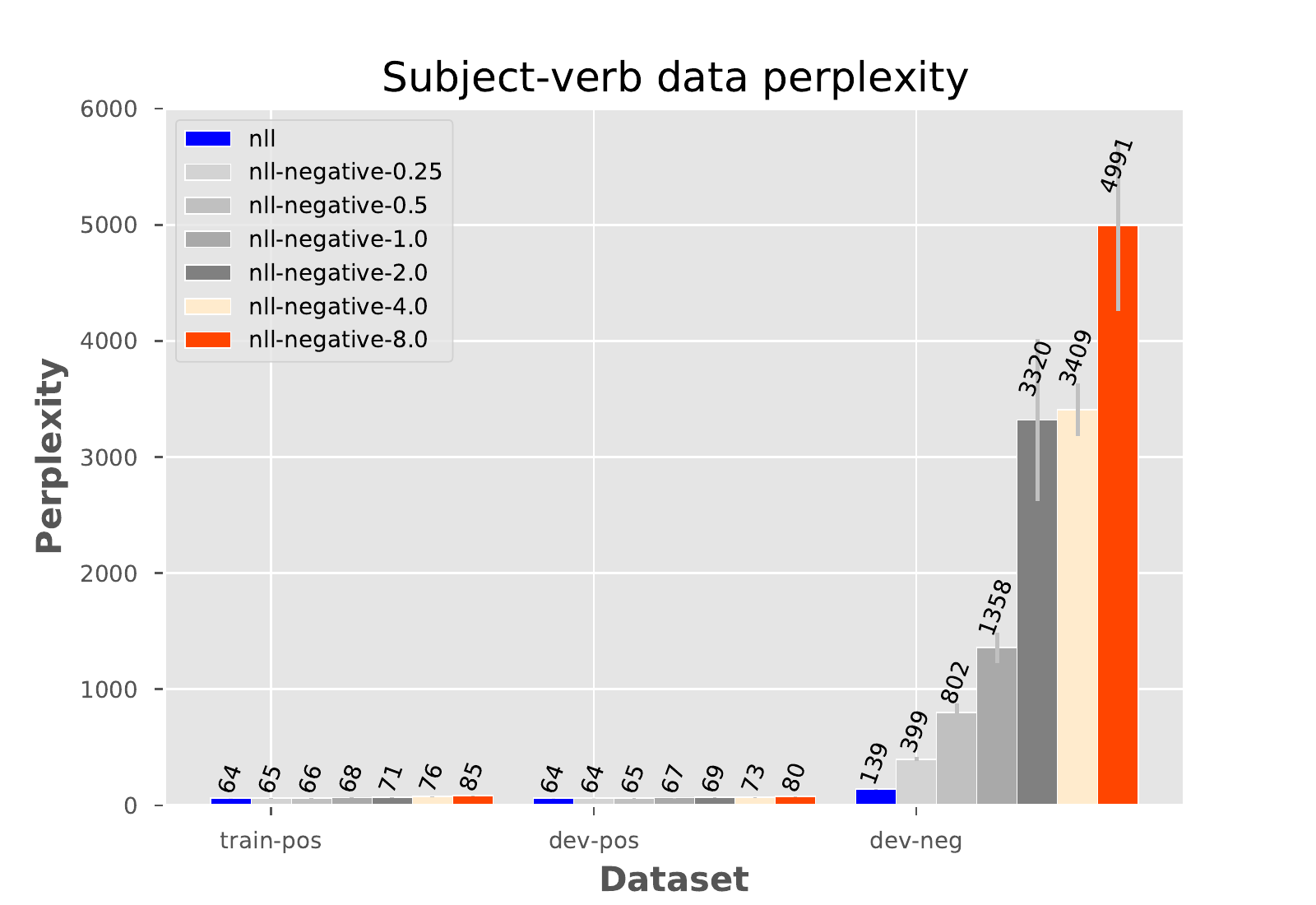}
\caption{Perplexity (LSTM large/1500h).}
   \label{fig:sva_ppl_lstm1500h}
 \end{subfigure}
        \begin{subfigure}[b]{0.48\textwidth}
   \includegraphics[width=\textwidth]{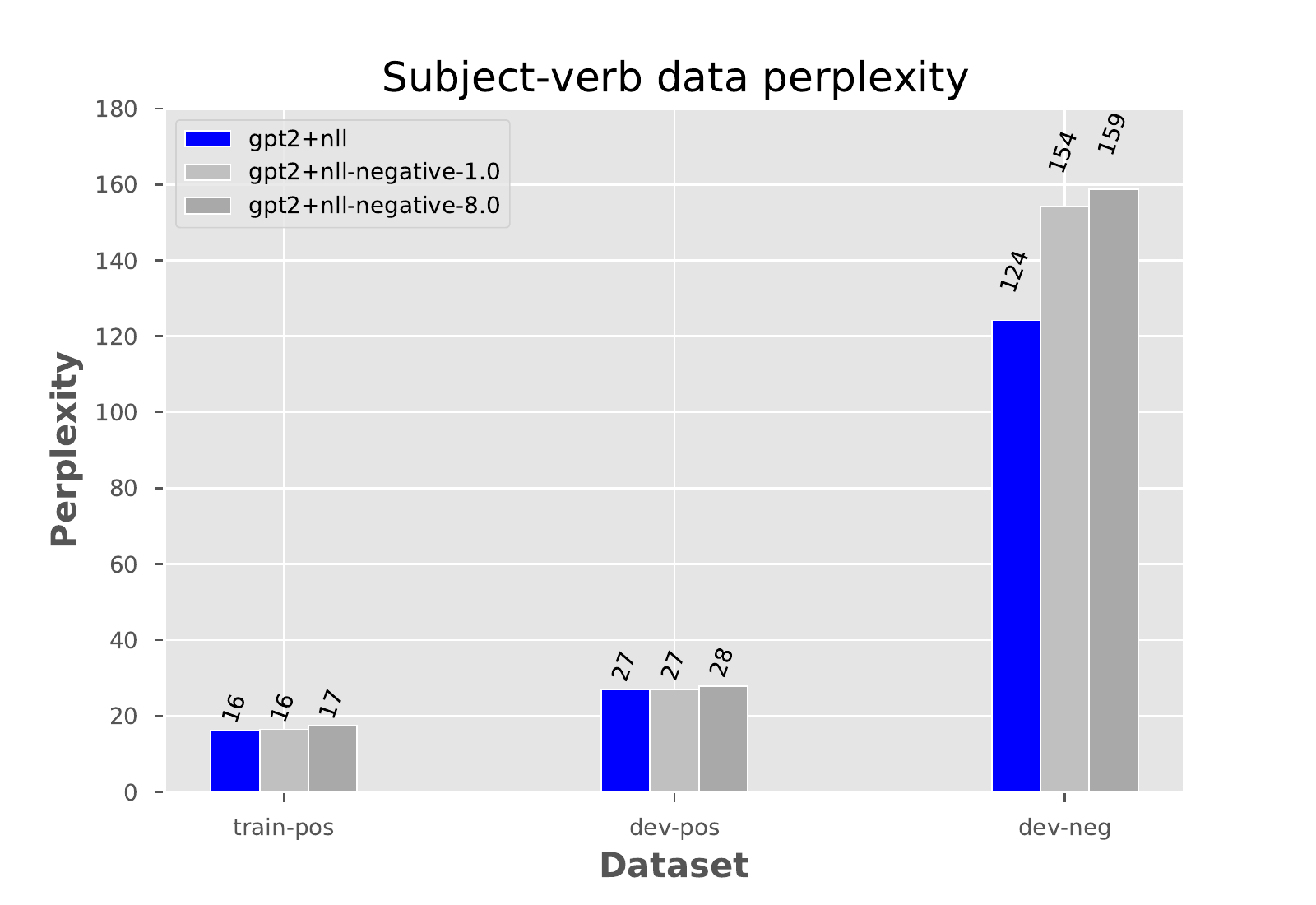}
\caption{Perplexity (GPT2).}
   \label{fig:sva_ppl_gpt2}
 \end{subfigure}
     \centering
   \begin{subfigure}[b]{0.48\textwidth}
   \includegraphics[width=\textwidth]{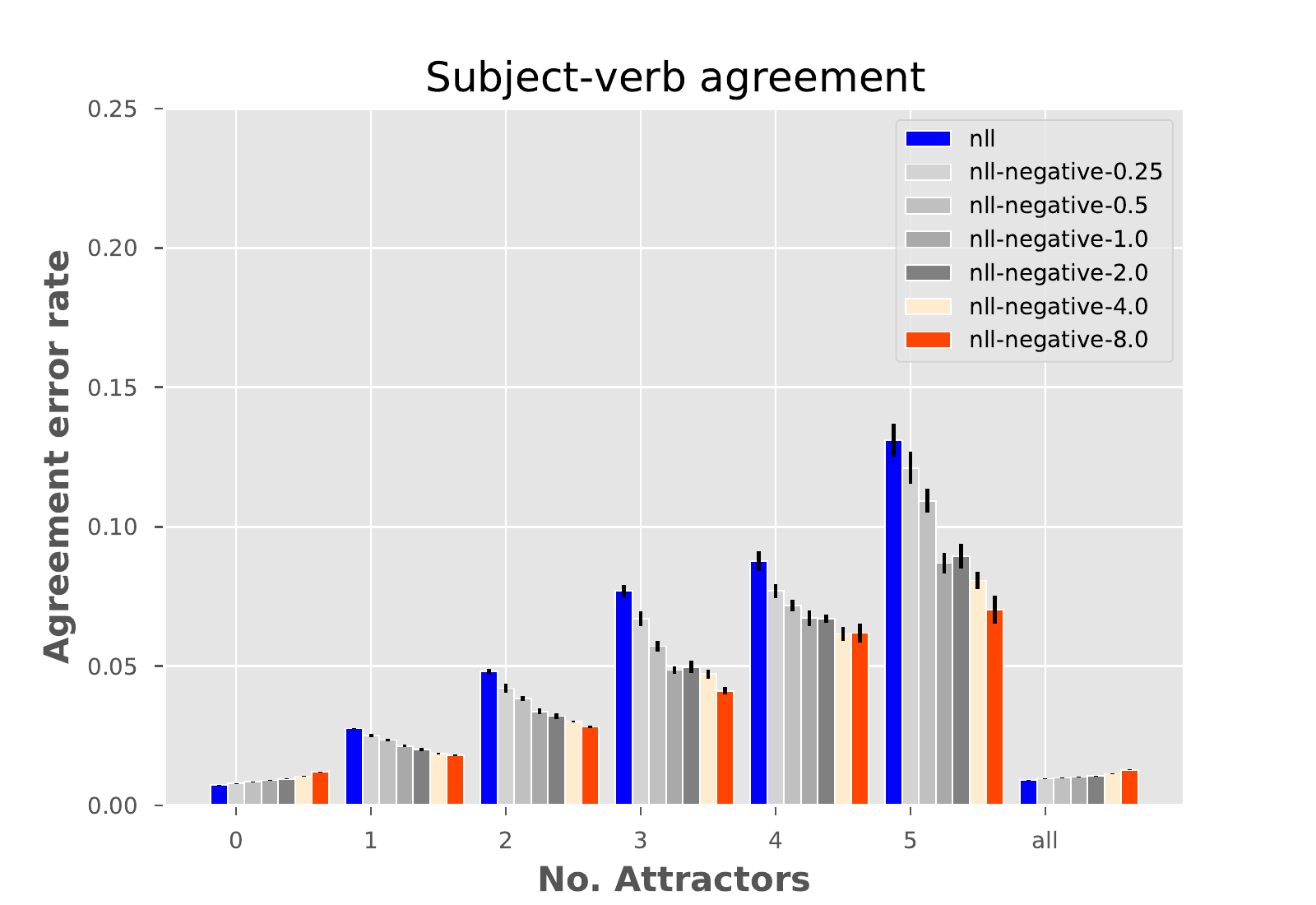}
\caption{Agreement error (LSTM large/1500h).}
   \label{fig:sva_lstm1500h}
   \end{subfigure}
  \begin{subfigure}[b]{0.48\textwidth}
   \includegraphics[width=\textwidth]{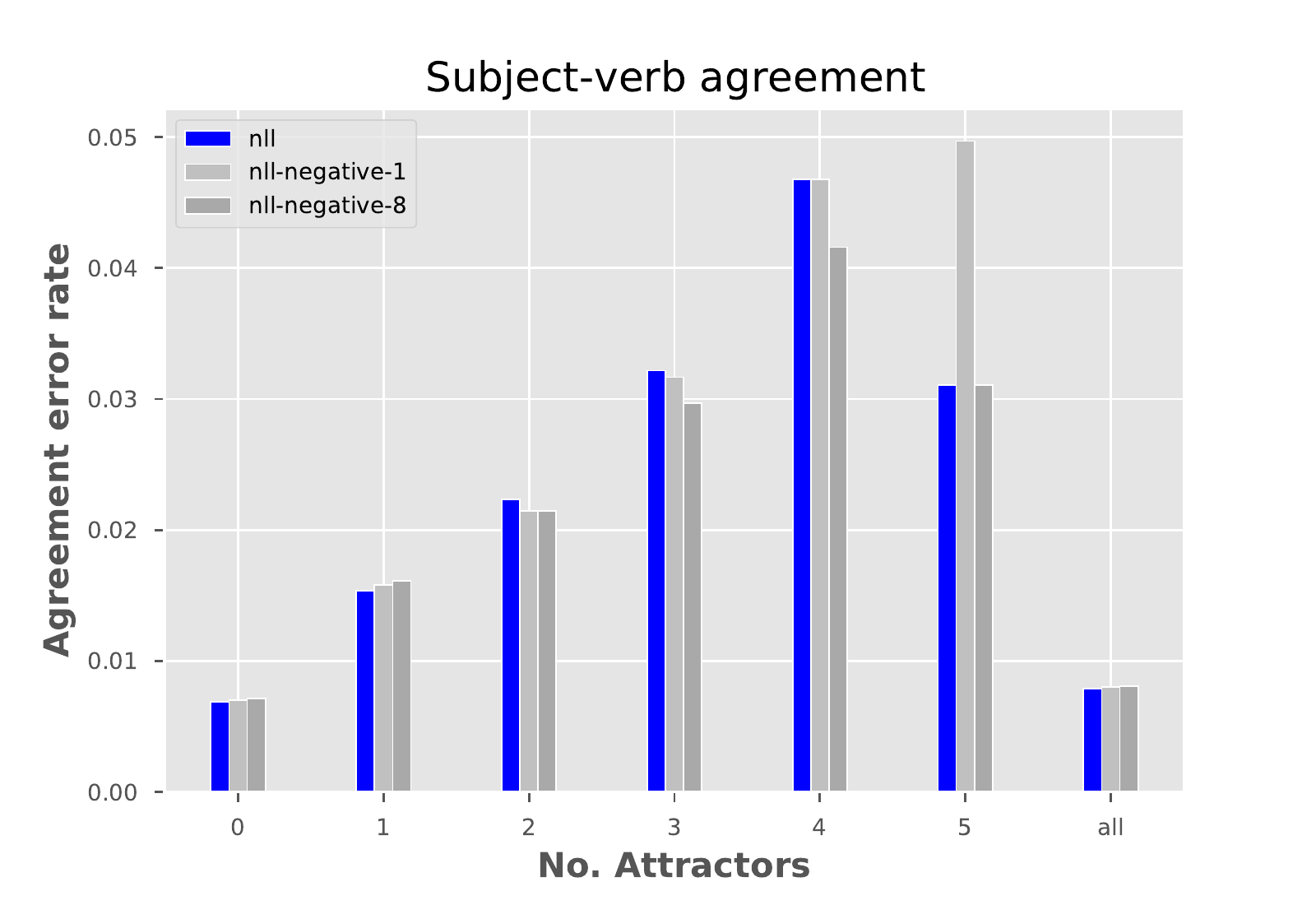}
\caption{Agreement error (GPT2).}
   \label{fig:sva_error_gpt2}
 \end{subfigure}
 \caption{Subject-verb agreement task with language models. Negative
   data are tri-gram generated.}
\label{fig:subject-verb}
 \end{figure}

 In the introduction, we gave an example of a sentence where n-gram
 statistics might prove misleading, because the verb that makes the sentence 
 ungrammatical is more common and co-occurs more frequently with the
 other words in the dataset.  It is unlikely that LSTMs have an inductive bias to overcome such
 signals, and indeed we saw in the previous experiment that they learn
 n-gram statistics.  We also saw that we can
 successfully attenuate these statistics with negative data, but does
 this cause the model to favor more desirable syntactic signals? 
 We now turn to a subject-verb agreement task \cite{linzen16assessing} to find evidence for the hypothesis
that removing n-grams allows the model to handle longer-distance syntactic dependencies.

The subject verb agreement task is to determine if the subject of a sentence
agrees with the verb.  For example, in “the keys are on the table,” the subject
“keys” must agree with the verb “are” in number, hence, “the keys is on the
table” is incorrect \cite{linzen16assessing}.  In general there might be an
intervening phrase between “keys” and “are” making the problem challenging as
in “The keys to the cabinet are on the table.”  Here, the singular noun
cabinet, which has an opposite number to “keys”, is an attractor which could
confuse the model.  There might be arbitrarily many of these intervening
attractors and the subject and verb may in general be arbitrarily far apart.
The results are thus organized by the number of attractors between the subject
and the verb.  The more attractors, the more difficult the task.  The language
model has correctly predicted agreement if it assigns higher probability to the
(grammatical) sentence with the correct verb than the (ungrammatical) sentence
with the incorrect verb.\footnote{This is a departure from previous work using
    language modeling to perform subject-verb agreement, which only considered
    the first part of the sentence up to the verb
\cite{linzen16assessing,kuncoro18lstms}.  We consider the entire sentence
because we're interested in language modeling and the verb can affect the
probability of downstream tokens.}

\begin{table}
    \caption{Error-rate comparison on subject-verb agreement data.}
    \label{tab:sva}
    \centering
    \begin{tabular}{|l|r|r|r|r|r|r|}
        \hline
        method & attr=0 & 1 & 2 & 3 & 4 & 5 \\
        \hline
        \hline
    	LSTM/nll (1500h) & \textbf{0.7} & 2.8 & 4.7 & 7.3 & 8.9 & 12.9 \\
    	LSTM/nll-neg-8 (1500h) & 1.2  & \textbf{1.8} & \textbf{2.8} & \textbf{4.1} & \textbf{6.2} & \textbf{7.0} \\
	\hline
	$\Delta$ & 0.5 & -1.0 & -1.9 & -3.2 & -2.7 & -5.9 \\
   	\hline
	\hline
	GPT2 (110M, zero-shot) & 7.7 & 14.8 & 21.1 & 24.9 & 21.5 & 24.2 \\
	GPT2+nll & \textbf{0.7} & \textbf{1.5} & 2.2 & 3.2 & 4.7 & \textbf{3.1} \\
	GPT2+nll-neg-8 & 0.7 & 1.6 & \textbf{2.1} & \textbf{3.0} & \textbf{4.2} & \textbf{3.1} \\
	\hline
	$\Delta$ (nll, nll-neg-8) & 0.0 & 0.1 & -0.1 & -0.2 & -0.5 & 0.0 \\
        \hline
    \end{tabular}
\end{table}

We employ the same splitting strategy as prior work ({\em ibid}), and train the
small and large LSTMs for 10 and 20 epochs respectively.  We also fine-tune and
evaluate a pretrained GPT2 model on the same splits (see
Appendix~\ref{app:gpt2-details} for fine-tuning details). 

Table~\ref{tab:sva} shows that, in general, training or fine-tuning by
attenuating n-gram statistics improves language models' ability to
associate verbs with their correct subject. As we increase the weight
on the negative data objective, model perplexity on the negative
development data increases (Figures \ref{fig:sva_ppl_lstm1500h} \&
\ref{fig:sva_ppl_gpt2}) and for the LSTM, the subject-verb agreement
error decreases (Figure~\ref{fig:sva_lstm1500h}), though this trend is
mixed for GPT-2 (Figure~\ref{fig:sva_error_gpt2}).  For the LSTM, this
performance gap increases as the number of attractors increases while
for GPT-2 the performance gap is roughly constant with most benefit at
four attractors.  It is also interesting to note that when there are
no attractors, the negative data actually hurts LSTM performance on
the task.  The reason is that without any attractors, the subject and
verb are close and the tri-grams serve as a good proxy for syntax,
despite the flaws we highlighted.

\section{Conclusion}
We proposed a method for detecting and attenuating undesirable signals
in language models.  When applied to detecting n-grams, we found that
existing models rely on them possibly at the expense of more syntactic
signals.  When applied to attenuating n-grams at train-time, our
method added an inductive bias that succesfully pushed the more
powerful language models (LSTMs, GPT2) away from the less powerful
models (tri-grams) and towards hypothesis classes that performed better
at syntactic tasks.  Perplexity was barely affected. The method is a
general way, like data augmentation, of imbuing the model with
inductive bias.  Indeed, while we focused
on n-grams in this work, the method is much more general and we can
apply it as a tool to detect and attenuate other undesirable signals
in the data, like gender bias. In future work, we also want to
explore ways of incorporating the idea into masked language models,
and explore alternatives to negative data like contrastive learning,
that more directly incorporate the negative model into the loss
function.

\clearpage

\bibliography{ref}
\bibliographystyle{plain}

\newpage
\appendix

\section{Additional experiments and details for the subject-verb data}

\begin{figure}
  \tikzstyle{data}=[draw, fill=blue!20, minimum size=2em]
\tikzstyle{model}=[draw, fill=blue!20, minimum size=2em]
\tikzstyle{positive}=[draw, fill=green!20, minimum size=2em]
\tikzstyle{negative}=[draw, fill=red!20, minimum size=2em]

\begin{tikzpicture}[node distance=4.5cm,auto,>=latex']
    \node [positive] (dpos) {positive data $\mathcal{D}$};
    \node [negative, right of=dpos] (mneg) {negative model $\mathcal{M}^\prime$};
    \node[negative, right of=mneg](dneg){negative data $\mathcal{D}^\prime$};
    \node[positive, right of=dneg](mpos){$\mathcal{M}$};
    \path[->] (dpos) edge node[below] {train $\mathcal{M}^\prime$} (mneg);
    \path[->] (mneg) edge node[below] {gen $\mathcal{D}^\prime$} (dneg);
    \path[->] (dneg) edge node[below] {{\em un}train $\mathcal{M}$} (mpos);
    \path[->] (dpos) edge[bend left,out=15, in=165] node[below] {train $\mathcal{M}$} (mpos);
\end{tikzpicture}
  \caption{A block diagram of the anti-modeling training procedure.  A
    negative generative model $\mathcal{M}^\prime$ is automatically
    fit to the positive data $\mathcal{D}$ and then used to generate
    negative data $\mathcal{D}^\prime$.  Both positive and
    negative data can then be used in the appropriate way to train the final model
    $\mathcal{M}$.  Red boxes indicate negative models and data, green boxes indicate
    positive model and data.}
  \label{fig:app-exor-diagram}
\end{figure}
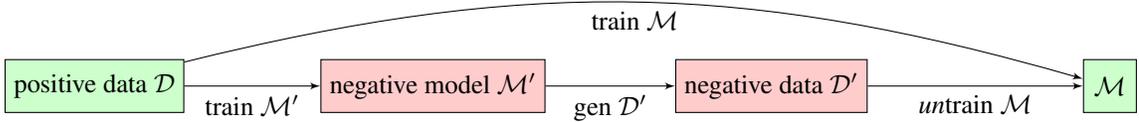

\subsection{Subject-verb data}
\label{app:data}
Here we provide more details of the subject-verb data.  We employ the same split of the data as prior work as best we can, based on their code, using the same random splitting strategy with the same random seed and the same proportions of train, test and dev \cite{linzen16assessing}.  We then organized the data into the number of intervening attractors, achieving data statistics very close, but not exactly identical to another piece of prior work \cite{kuncoro18lstms}.  We report our data statistics for the testing set in Table~\ref{tab:data-sva}, which we can compare with Table 1 in prior work to see the similarity in our data statistics ({\em ibid}).  We report the \% of instances that each set represents as a percentage of the set of data with opposite number attractors and as a percentage of the full test set (which is larger because it includes cases in which there are attractors of the same number or cases with more than 5 attractors).

\begin{table}
  \caption{Subject-verb agreement data statistics.}
  \label{tab:data-sva}
  \centering
  \begin{tabular}{r|r|r|r}
    \# Attractors & \# Instances & \% Instances ($n=1 \ldots 5$) & \% Instances (full test)\\
    \hline
    $n=0$ & 1,146,256 & 94.65\% & 80.75\% \\
    $n=1$ & 51,714 & 4.34\% & 3.71\% \\
    $n=2$ & 9,151 & 0.78\% & 0.66\% \\
    $n=3$ & 1,942 & 0.17\% & 0.14\% \\
    $n=4$ & 546 & 0.05\% & 0.04\% \\
    $n=5$ & 161 & <0.01\% & <0.01\% \\
    \hline
    $n\in\{0\ldots5\}$ & 1,211,033 & 100\% & 85\% \\
    all & 1,419,491 &  & 100\% \\
    \hline
  \end{tabular}
\end{table}

\subsection{Subject-verb agreement comparison with related work}
We note that our goal is not to achieve state of the art on the
subject-verb agreement task, but rather our goal is to use this task
to help provide evidence for our hypothesis: attenuating n-gram
signals improves syntactic ability.  Nevertheless, it might be useful
to compare how our language models perform with other results reported
in the literature on this task, which we do in Table~\ref{tab:related}.

We caution that due to various factors, and different practices
employed throughout the literature, that these results are not directly
comparable.  We attempt a comprehensive set of differences here:
\begin{itemize}
  \item We compute the probability of the full sentences
(grammatical and ungrammatical) when making subject-verb agreement
judgements~\cite{marvin2018targeted,goldberg19assessing}, but other work only computes
the probability of the tokens up to the
verb~\cite{linzen16assessing,kuncoro18lstms}.

\item BERT and BART models are trained on Wikipedia and hence are
  exposed to the sentences in the test set of the subject-verb
  agreement (which are all from Wikipedia also).

\item BERT and BART models employ a different tokenization scheme that
  can split words into multiple pieces.  Sentences for which the
  target verb is split into multiple pieces are filtered out of the
  BERT evaluation data.  If it happens to be the case that these words
  are relatively rarer, then it might mean that this subset of the
  data is slightly easier. 

\item BERT and BART models are evaluated on a subset of the evaluation
  data.  Sentences in which the main verb are is/are are\footnote{[are is are are]
  must be an exceedingly rare 4-gram!} pruned because in copula
constructions, the objects after the verb can provide syntactic cues
for this task (e.g., the object ``friends'' has the same number as
``girls'' in the sentence ``the girls are friends.''); however, note
that the example sentence we employ in the introduction is also
copular and this is a case in which the object following the verb
functions as a {\em distractor} in the sense that the number of the
object disagrees with the subject.  Thus, it is unclear how this
pruning might affect the results.
\end{itemize}

If we compare the first two rows of Table~\ref{tab:related}, we see
that the negative data training method allows a model with just 200 hidden units (8M
parameters) to close the gap with the 1500 hidden-unit LSTM (69M parameters) in
some categories; though, performance is still worse.  If we look at
the negative data training method on the 1500 hidden-unit LSTM, we see
that it greatly improves over the same LSTM trained with positive data
only (with the exception of the 0 attractor case, which we discussed
earlier in the experiments section).

When we compare with other results reported in the literature, such as
recursive neural network grammars (RRNG-$\star$), that also add an
inductive bias to language, we see that our method has a substantially
lower error rate.  However, our method employs the entire sentence to
judge grammaticallity and their method only employs up to the verb.
Moreover, it is unclear how many hidden units their model employs and
it is likely that much of our performance increase is simply because
our 1500 hidden unit LSTM is larger.

Now interestingly, our 84M parameter LSTM trained with negative data
(third row) performs much better than BART, a transformer-based
denoising auto-encoder, that has 400M parameters; and performs
comparably to BERT (300M parameters).  This is remarkable especially
since the training data of BART and BERT include all the testing
sentences in this task, and these massive models dwarf our relatively
modest LSTM. 

\subsection{Other negative n-gram distributions: bi-grams and 4-grams}
\label{app:n-gram}
For our experiments in the main paper we chose word tri-grams as our
negative distribution as we suspected them to be a sweet spot for our
dataset: they are powerful enough to generate reasonable language-like
sentences, but not powerful enough to overfit the positive training
data (and thus generate negative data that is too similar).  Our
results for the 2-grams and 4-gram negative models still demonstrate
some benefits, but they are not as pronounced or uniform as for the
tri-gram models.

\begin{figure}
   \centering
  \begin{subfigure}[b]{0.49\textwidth}
   \includegraphics[width=\textwidth]{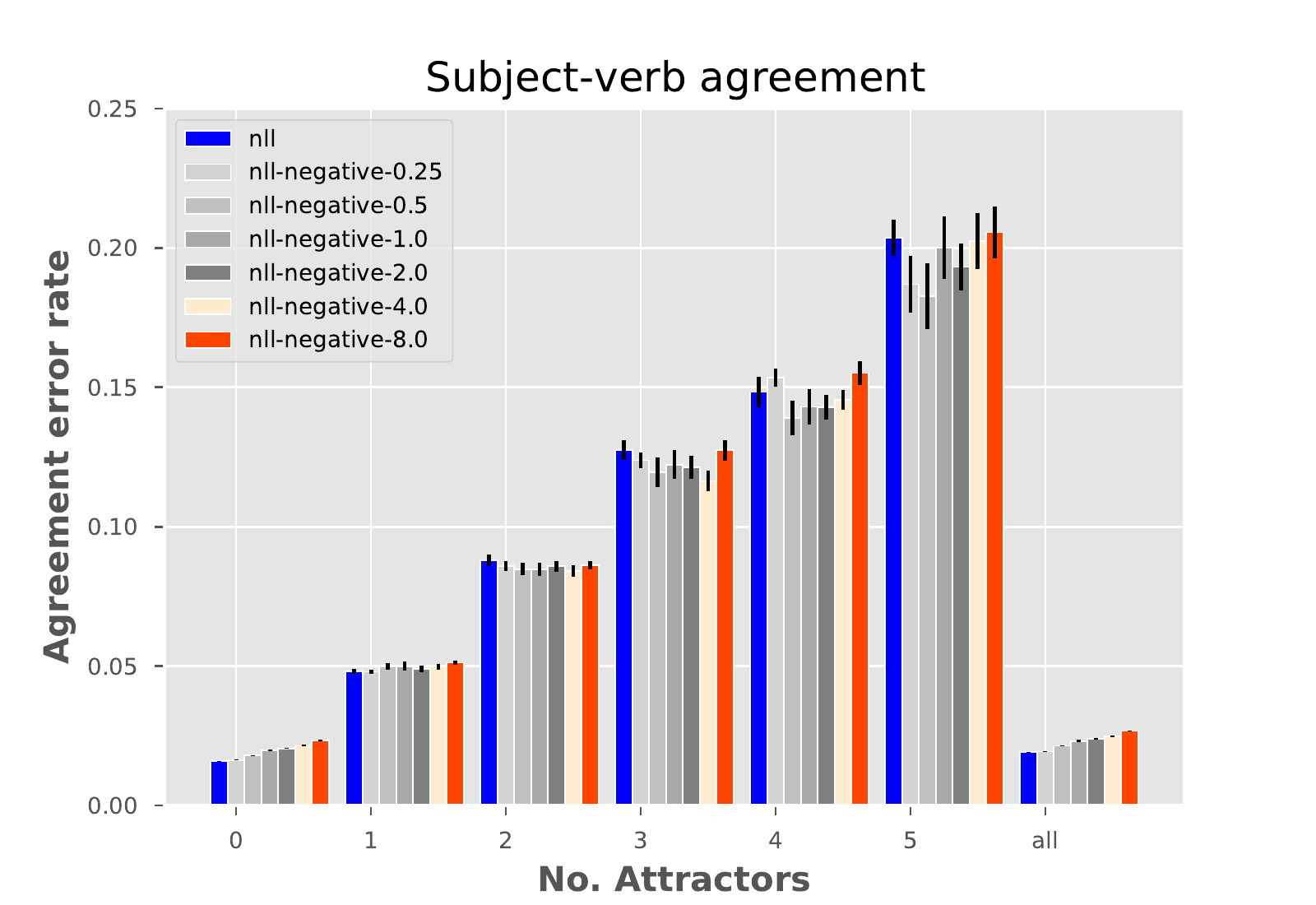}
\caption{Agreement error (negative: bi-grams).}
   \label{fig:sva_lstm200h-2g}
   \end{subfigure}
   \centering
  \begin{subfigure}[b]{0.48\textwidth}
   \includegraphics[width=\textwidth]{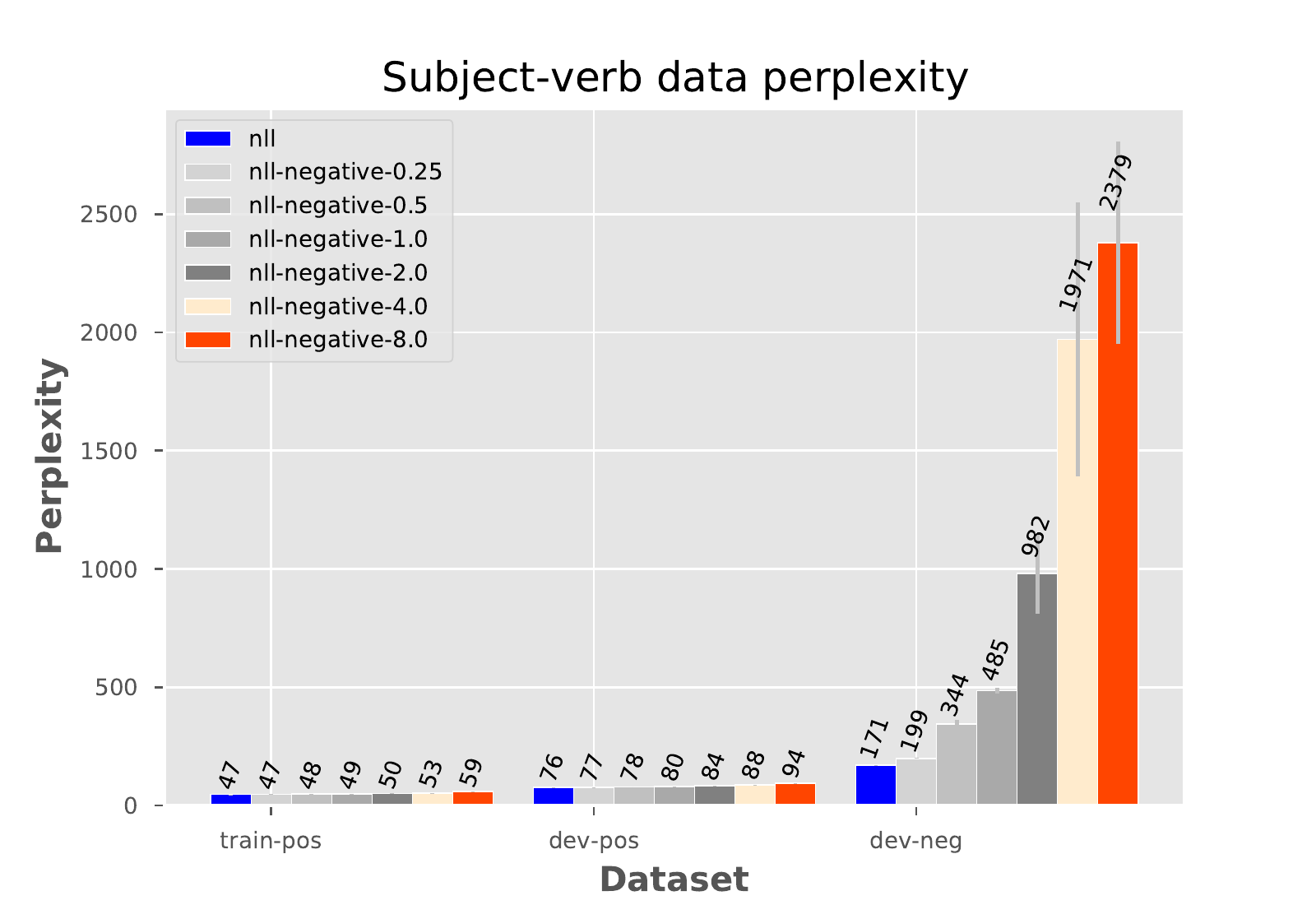}
\caption{Agreement data ppl (negative: bi-grams)}
   \label{fig:sva_ppl_lstm200h-2g}
 \end{subfigure}
  \begin{subfigure}[b]{0.49\textwidth}
   \includegraphics[width=\textwidth]{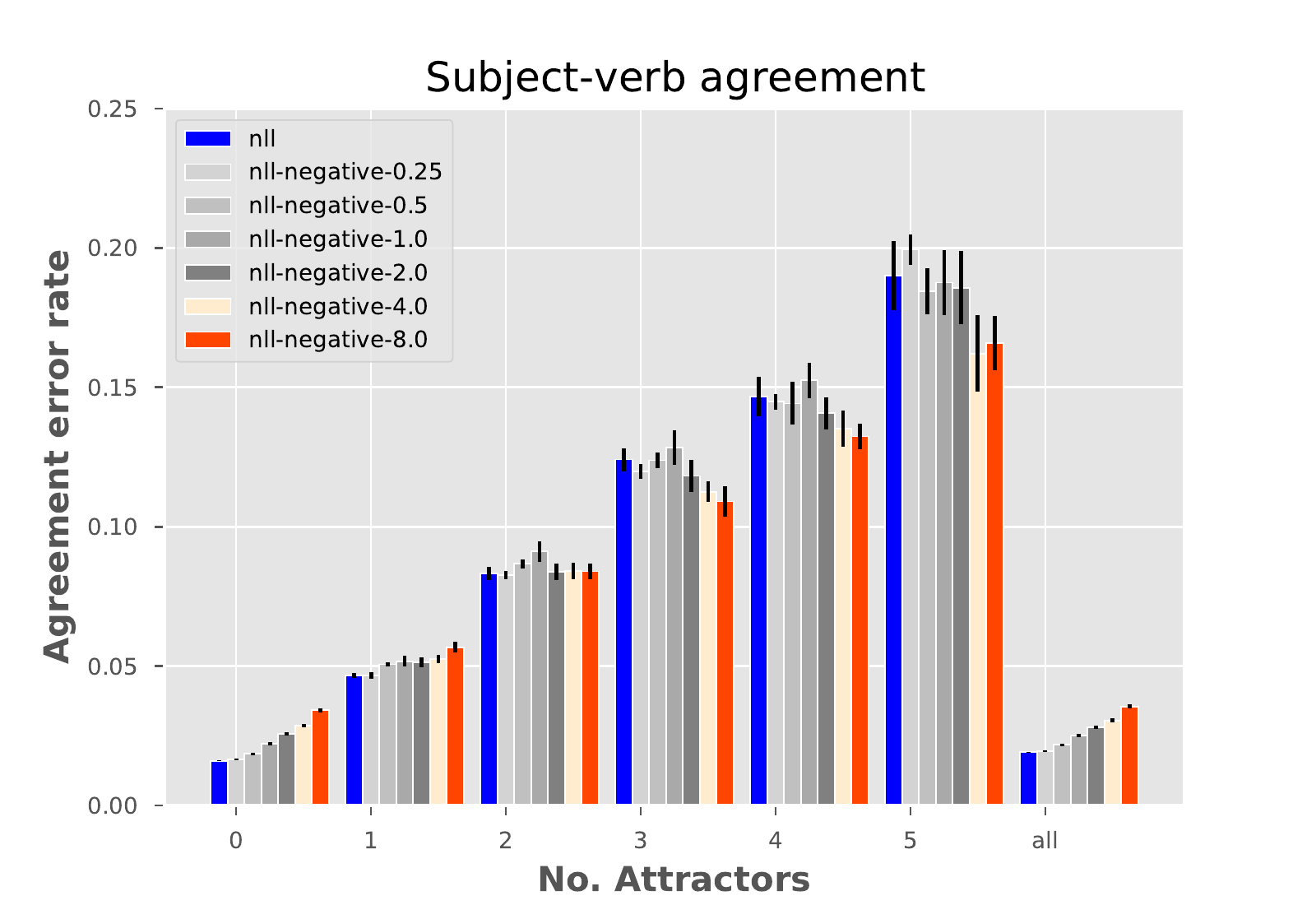}
\caption{Agreement error (negative: 4-grams).}
   \label{fig:sva_lstm200h-4g}
   \end{subfigure}
   \centering
  \begin{subfigure}[b]{0.48\textwidth}
   \includegraphics[width=\textwidth]{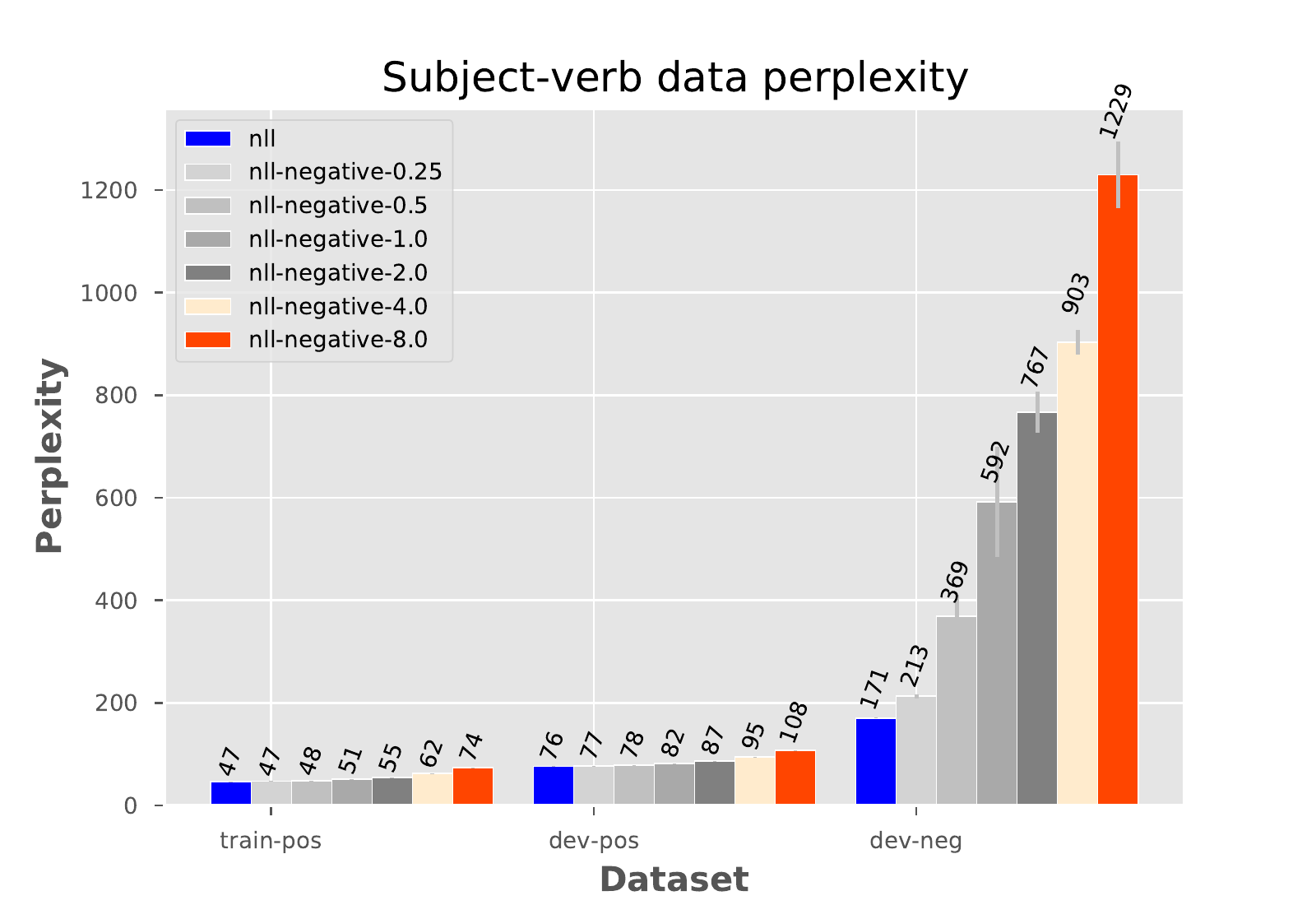}
\caption{Agreement data ppl (negative: 4-grams).}
   \label{fig:sva_ppl_lstm200h-4g}
 \end{subfigure}
 \caption{Subject-verb agreement data with the 200 hidden unit LSTM.}
 \label{fig:sva_lstm200h-ng}
\end{figure}

\begin{figure}
   \centering
  \begin{subfigure}[b]{0.49\textwidth}
   \includegraphics[width=\textwidth]{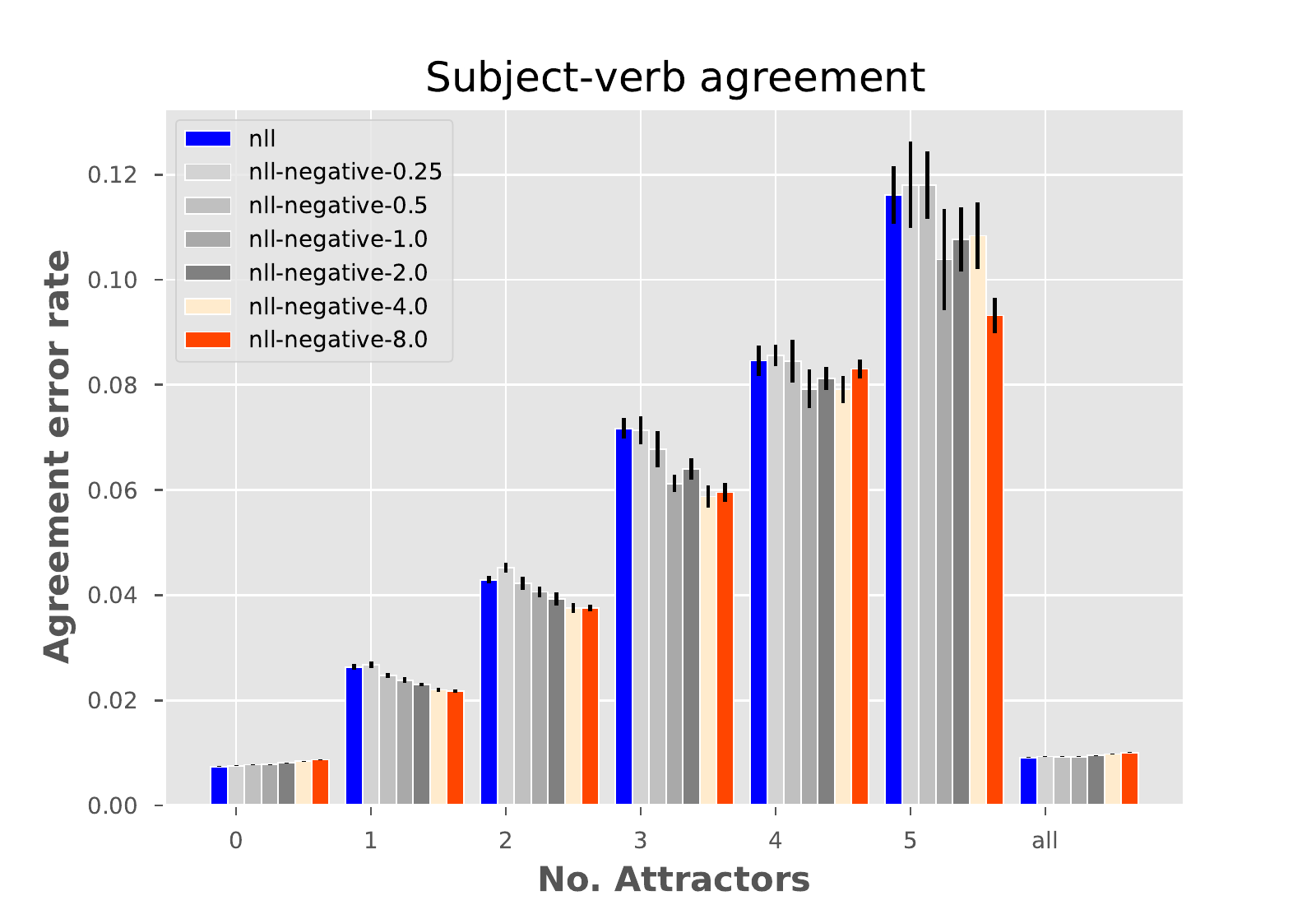}
\caption{Agreement error (negative: bi-grams).}
   \label{fig:sva_lstm1500h-2g}
   \end{subfigure}
   \centering
  \begin{subfigure}[b]{0.48\textwidth}
   \includegraphics[width=\textwidth]{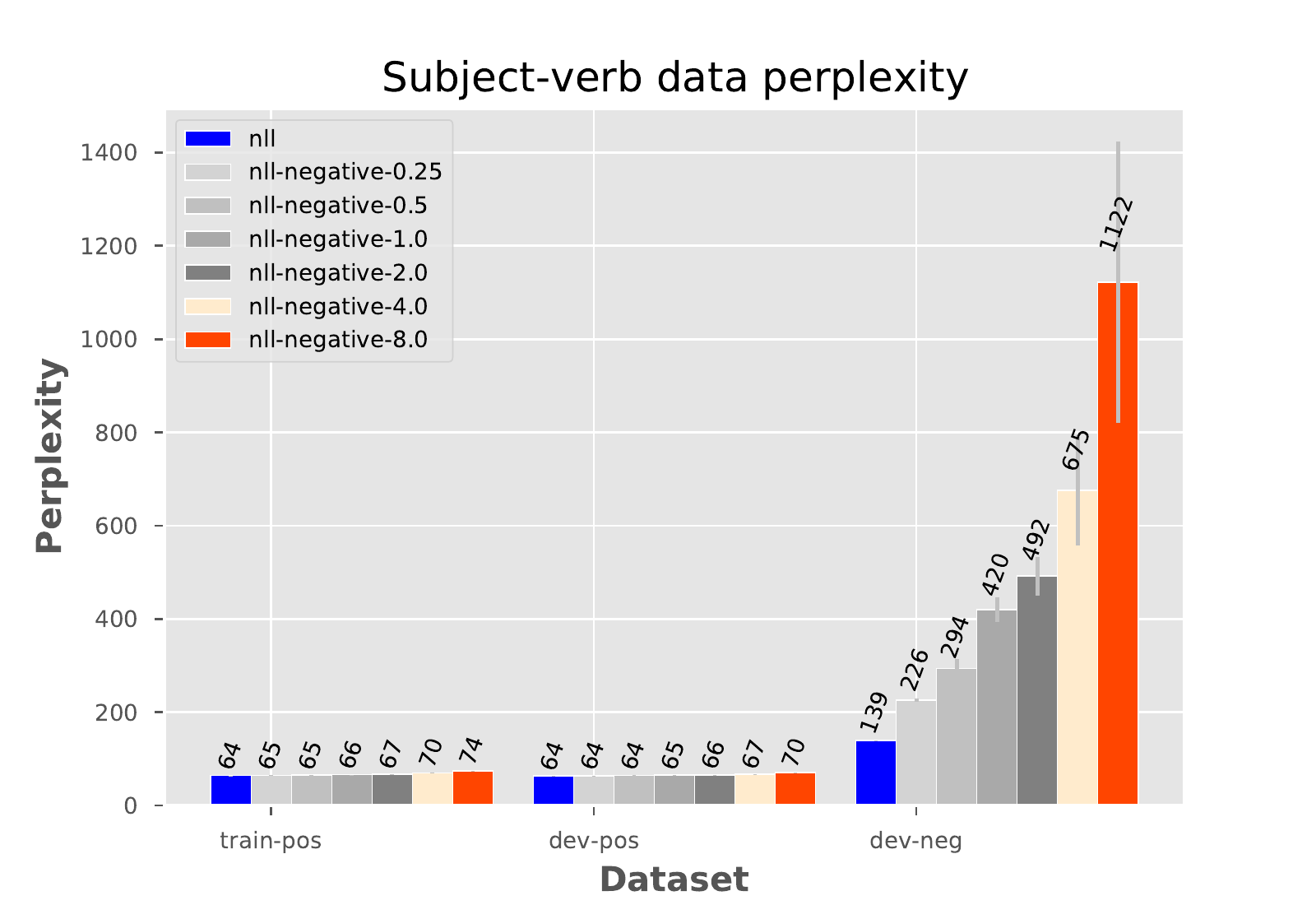}
\caption{Agreement data ppl (negative: bi-grams)}
   \label{fig:sva_ppl_lstm1500h-2g}
 \end{subfigure}
  \begin{subfigure}[b]{0.49\textwidth}
   \includegraphics[width=\textwidth]{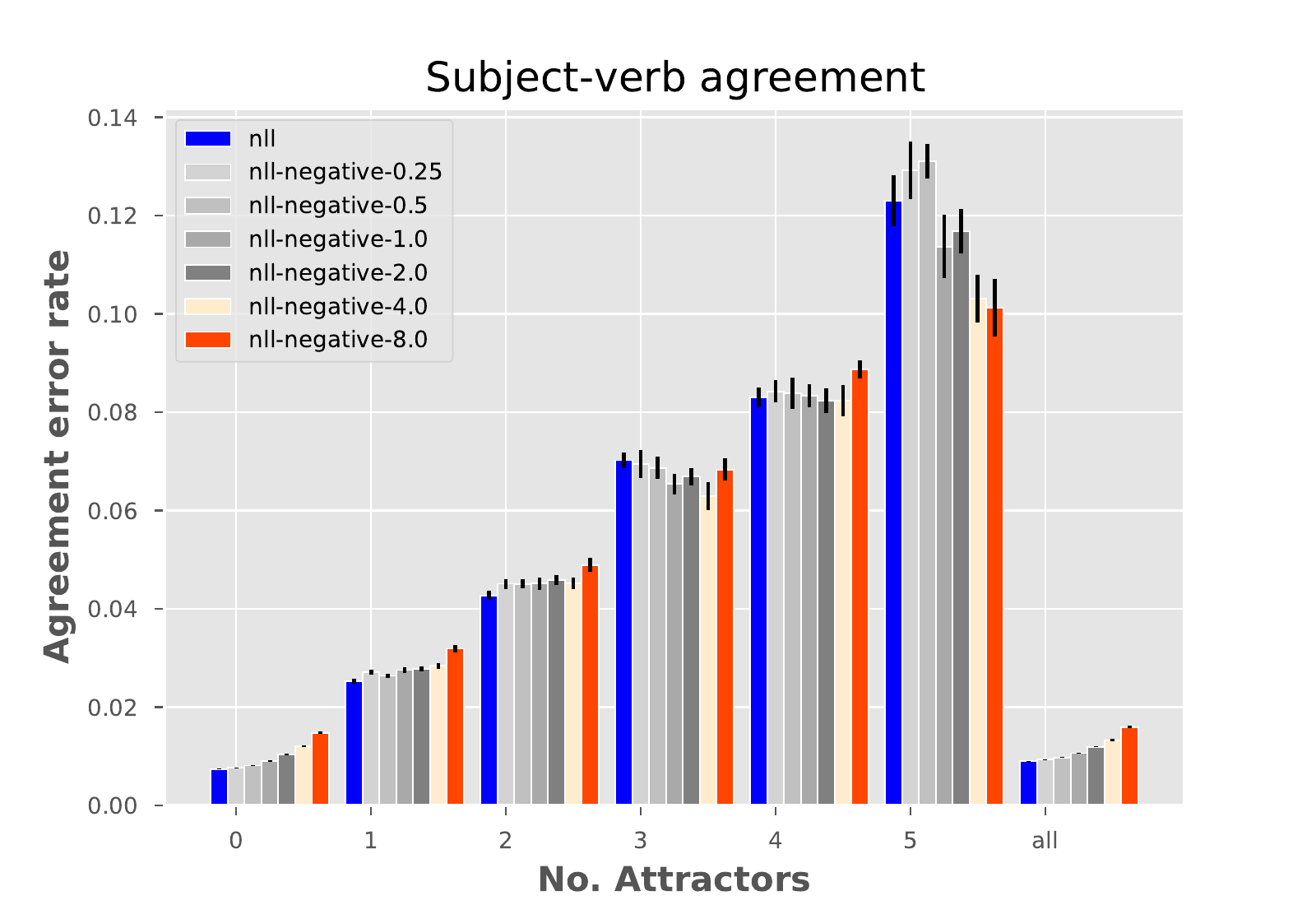}
\caption{Agreement error (negative: 4-grams).}
   \label{fig:sva_lstm1500h-4g}
   \end{subfigure}
   \centering
  \begin{subfigure}[b]{0.48\textwidth}
   \includegraphics[width=\textwidth]{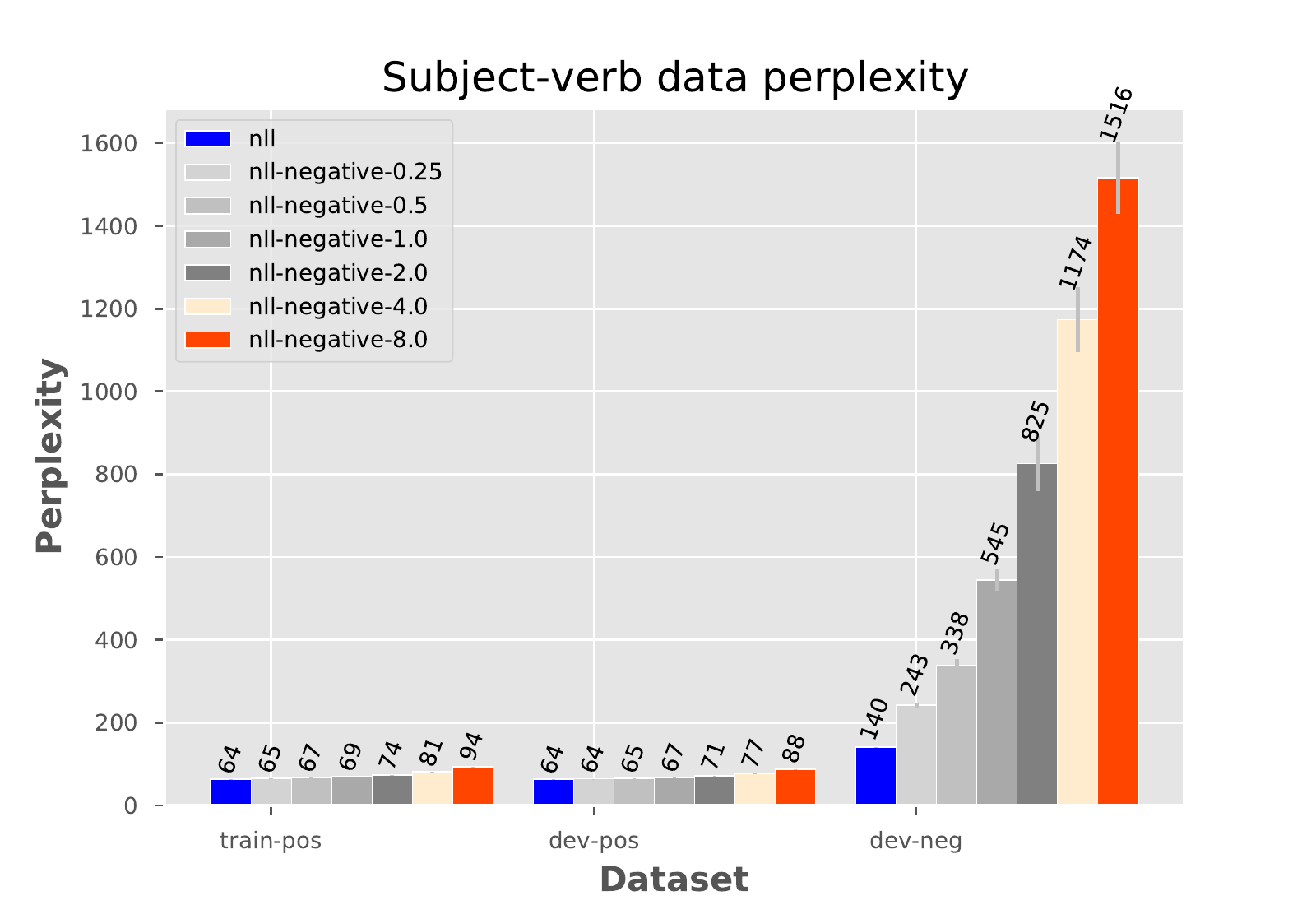}
\caption{Agreement data ppl (negative: 4-grams).}
   \label{fig:sva_ppl_lstm1500h-4g}
 \end{subfigure}
 \caption{Subject-verb agreement data with the 1500 hidden unit LSTM.}
 \label{fig:sva_lstm1500h-ng}
 \end{figure}

We present results for the small 200 hidden unit LSTM
in Figure~\ref{fig:sva_lstm200h-ng}.  We present results for the large 1500
hidden unit LSTM in Figure~\ref{fig:sva_lstm1500h-ng}.  The results confirm
our suspicions that a 4-gram is
too strong of an anti-model for the relatively small train-set of the
subject-verb dataset (most of the data is relegated to the test-set in
order to allow more accurate analysis).  One indication of this is that
for the 1-attractor, the models that employs the 4-grams for
negative data performs worse than the model that only employs positive
data, which was not the case for tri-grams.  For the larger
LSTM, we also see that 4-grams are problematic for some larger
attractor conditions too.  We suspect that 4-grams
are overfitting the training set and that the sentences they generate
are too similar to the positive sentences after which they are
modeled.  So using a negative model that is too powerful is not ideal.

On the other hand, using a model that is too weak, like bi-grams, is
not effective as using a more powerful negative model, like tri-grams
(this is likely to be particular to the subject-verb training set, and
others like it that are relatively small, so one should not generalize
too far here).  For the smaller LSTM, there is only an improvement for
smaller values of $\alpha$ and the improvement is not nearly as
pronounced as in the tri-gram version.  For the larger LSTM, there are
improvements across all values of $\alpha$, but the improvement again
is not as large as in the tri-gram version.

\begin{table}[h]
  \centering
  \begin{tabular}{|l|r|r|r|r|r|r|}
    \hline
    method & attr=0 & 1 & 2 & 3 & 4 & 5 \\
    \hline
    nll (1500h) & 0.7 & 2.8 & 4.7 & 7.3 & 8.9 & 12.9 \\
    nll-neg-8.0 (200h) & 2.8 & 4.2 & 6.5 & 8.9 & 10.2 & 11.7 \\
    \hline
    nll-neg-8.0 (1500h) & 1.2  & 1.8 & 2.8 & 4.1 & 6.2 & 7.0 \\
    \hline
    RNNG-TD \cite{kuncoro18lstms} & & & 5.5 & 7.8 & 8.9 & \\
    RNNG-LC \cite{kuncoro18lstms} & & & 5.4 & 8.2 & 9.9 & \\
    RNNG-BU \cite{kuncoro18lstms} & & & 5.7 & 8.5  & 9.7 & \\
    \hline
    LSTM 50h \cite{linzen16assessing,kuncoro18lstms} &2.4 & 8.0 & 15.7& 26.1& 34.6 & \\
    LSTM 150h \cite{kuncoro18lstms} & 1.5 & 4.5 & 9.0 & 14.3 & 17.6 & \\
    LSTM 250h \cite{kuncoro18lstms} & 1.4 & 3.3 & 5.9 & 9.7 & 13.9 & \\
    LSTM 350h \cite{kuncoro18lstms} & 1.3 & 3.0 & 5.7 & 9.7 & 13.8 & \\
    1B Word LSTM \cite{jozefowicz2016exploring,kuncoro18lstms} & 2.8 & 8.0 & 14.0 & 21.8 & 20.0 & \\
       \hline
       BERT (340M params)*
       \cite{goldberg19assessing,devlin2019bert} & & 3\hphantom{.0} & 3\hphantom{.0} & 4\hphantom{.0} &
                                                                       3\hphantom{.0} & \\
       BART (400M params)*\cite{jafer2020assessing,lewis2019bart} &
                    & 3.8 & 4.1 & 6.0 & 6.7 & \\
       GPT (110M params) \cite{radford18improving,wolf19some} &
                    & 18\hphantom{.0} & 24\hphantom{.0} & 31\hphantom{.0} & 30\hphantom{.0} & \\
    \hline
     \end{tabular}
     \caption{Error-rate comparison to other language models on the subject-verb
       agreement data.  Note these results are not directly
       comparable for reasons stated in the text: in short, BERT and
       BART are trained on data that includes the test sentences and
       are also evaluated on a subset of the official test-set.  The
       other LSTMs and RGNN models only model sentences up to the main
       verb. *trainining set includes test data.}
     \label{tab:related}
 \end{table}

 \section{Results on targeted syntactic evaluation}

 We also evaluate our models on an additional syntactic tasks with
 artificial data from prior work that is automatically generated from
 hand-crafted grammars \cite{marvin2018targeted}.  Since the models we
 study in this work are trained on the subject-verb agreement task
 \cite{linzen16assessing}, which is an order of magnitude smaller than
 the training data of prior work (ibid), it turns out that more than
two-thirds of the data has an example with an out-of-vocabulary (OOV)
 word, and that in 24\% of the cases, this out of vocabulary word
 happened to be the word (the {\em target} word) that distinguishes
 the grammatical version of the sentence from the ungrammatical
 version.  We present three sets of results, one on the entire dataset
 (Table~\ref{tab:marvin-linzen-all}), one on a subset comprising examples
 for which the target is not OOV
 (Table~\ref{tab:marvin-linzen-oov-target}) and one on a subset comprising
 examples for which no word is OOV (Table~\ref{tab:marvin-linzen-oov})
 . This latter is done because most words in these sentences are
 determiners and other fillers and thus the OOVs might be the cues
 relevant to perform the task.  Highlighted rows indicate those for
 which the negative data seemed to improve the baseline of negative
 log likelihood (nll) on the positive data only.

 \begin{table}
   \caption{LSTM 200h results on Marvin \& Linzen subset without oov targets.}
   \label{tab:marvin-linzen-oov-target}
  \begin{tabular}{lrrrrr} 
  \hline \\ 
 & \multicolumn{1}{l}{nll} & \multicolumn{1}{l}{nll-neg-1.0} & \multicolumn{1}{l}{nll-neg-2.0} & \multicolumn{1}{l}{nll-neg-4.0} & \multicolumn{1}{l}{nll-neg-8.0} \\ 
  \hline \\ 
\textsc{Subject-verb agreement} & \multicolumn{4}{c}{} \\ 
  Simple & $ 1.00$ \textcolor{gray}{$\pm 0.00$} & $ 1.00$ \textcolor{gray}{$\pm 0.00$} & $ 1.00$ \textcolor{gray}{$\pm 0.00$} & $ 0.92$ \textcolor{gray}{$\pm 0.03$} & $ 0.84$ \textcolor{gray}{$\pm 0.06$} \\ 
  In a sentential complement & $ 0.86$ \textcolor{gray}{$\pm 0.01$} & $ 0.85$ \textcolor{gray}{$\pm 0.02$} & $ 0.86$ \textcolor{gray}{$\pm 0.02$} & $ 0.77$ \textcolor{gray}{$\pm 0.03$} & $ 0.73$ \textcolor{gray}{$\pm 0.03$} \\ 
  \rowcolor{lightgray} Short VP coordination & $ 0.95$ \textcolor{gray}{$\pm 0.02$} & $ 0.99$ \textcolor{gray}{$\pm 0.00$} & $ 0.95$ \textcolor{gray}{$\pm 0.01$} & $ 0.77$ \textcolor{gray}{$\pm 0.05$} & $ 0.74$ \textcolor{gray}{$\pm 0.06$} \\ 
  \rowcolor{lightgray} Long VP coordination & $ 0.62$ \textcolor{gray}{$\pm 0.01$} & $ 0.66$ \textcolor{gray}{$\pm 0.04$} & $ 0.59$ \textcolor{gray}{$\pm 0.01$} & $ 0.59$ \textcolor{gray}{$\pm 0.03$} & $ 0.56$ \textcolor{gray}{$\pm 0.02$} \\ 
  Across a prepositional phrase & $ 0.96$ \textcolor{gray}{$\pm 0.00$} & $ 0.96$ \textcolor{gray}{$\pm 0.00$} & $ 0.92$ \textcolor{gray}{$\pm 0.03$} & $ 0.78$ \textcolor{gray}{$\pm 0.07$} & $ 0.74$ \textcolor{gray}{$\pm 0.07$} \\ 
  \rowcolor{lightgray} Across a subject relative clause & $ 0.95$ \textcolor{gray}{$\pm 0.02$} & $ 0.98$ \textcolor{gray}{$\pm 0.01$} & $ 0.95$ \textcolor{gray}{$\pm 0.02$} & $ 0.83$ \textcolor{gray}{$\pm 0.04$} & $ 0.77$ \textcolor{gray}{$\pm 0.07$} \\ 
  \rowcolor{lightgray} Across an object relative clause & $ 0.76$ \textcolor{gray}{$\pm 0.02$} & $ 0.79$ \textcolor{gray}{$\pm 0.02$} & $ 0.80$ \textcolor{gray}{$\pm 0.02$} & $ 0.65$ \textcolor{gray}{$\pm 0.04$} & $ 0.63$ \textcolor{gray}{$\pm 0.05$} \\ 
  \rowcolor{lightgray} Across an objective relative (no that) & $ 0.70$ \textcolor{gray}{$\pm 0.02$} & $ 0.72$ \textcolor{gray}{$\pm 0.02$} & $ 0.74$ \textcolor{gray}{$\pm 0.03$} & $ 0.62$ \textcolor{gray}{$\pm 0.04$} & $ 0.58$ \textcolor{gray}{$\pm 0.03$} \\ 
  In an object relative clause & $ 0.64$ \textcolor{gray}{$\pm 0.02$} & $ 0.64$ \textcolor{gray}{$\pm 0.01$} & $ 0.64$ \textcolor{gray}{$\pm 0.01$} & $ 0.62$ \textcolor{gray}{$\pm 0.01$} & $ 0.59$ \textcolor{gray}{$\pm 0.01$} \\ 
  In an object relative (no that) & $ 0.63$ \textcolor{gray}{$\pm 0.02$} & $ 0.60$ \textcolor{gray}{$\pm 0.01$} & $ 0.64$ \textcolor{gray}{$\pm 0.01$} & $ 0.60$ \textcolor{gray}{$\pm 0.01$} & $ 0.56$ \textcolor{gray}{$\pm 0.02$} \\ 
  \hline \\ 
\textsc{Reflexive anaphora} & \multicolumn{4}{c}{} \\ 
  Simple (RA) & $ 0.84$ \textcolor{gray}{$\pm 0.03$} & $ 0.79$ \textcolor{gray}{$\pm 0.04$} & $ 0.79$ \textcolor{gray}{$\pm 0.04$} & $ 0.65$ \textcolor{gray}{$\pm 0.05$} & $ 0.66$ \textcolor{gray}{$\pm 0.04$} \\ 
  In a sentential complement & $ 0.86$ \textcolor{gray}{$\pm 0.01$} & $ 0.85$ \textcolor{gray}{$\pm 0.02$} & $ 0.86$ \textcolor{gray}{$\pm 0.02$} & $ 0.77$ \textcolor{gray}{$\pm 0.03$} & $ 0.73$ \textcolor{gray}{$\pm 0.03$} \\ 
  Across a relative clause (RA) & $ 0.75$ \textcolor{gray}{$\pm 0.01$} & $ 0.71$ \textcolor{gray}{$\pm 0.03$} & $ 0.67$ \textcolor{gray}{$\pm 0.03$} & $ 0.67$ \textcolor{gray}{$\pm 0.04$} & $ 0.66$ \textcolor{gray}{$\pm 0.02$} \\ 
  \hline \\ 
\textsc{Negative polarity items} & \multicolumn{4}{c}{} \\ 
  \rowcolor{lightgray} Simple (NPI) & $ 0.53$ \textcolor{gray}{$\pm 0.01$} & $ 0.55$ \textcolor{gray}{$\pm 0.02$} & $ 0.53$ \textcolor{gray}{$\pm 0.02$} & $ 0.54$ \textcolor{gray}{$\pm 0.01$} & $ 0.49$ \textcolor{gray}{$\pm 0.02$} \\ 
  \rowcolor{lightgray} Across a relative clause (NPI) & $ 0.56$ \textcolor{gray}{$\pm 0.01$} & $ 0.57$ \textcolor{gray}{$\pm 0.01$} & $ 0.57$ \textcolor{gray}{$\pm 0.01$} & $ 0.54$ \textcolor{gray}{$\pm 0.01$} & $ 0.54$ \textcolor{gray}{$\pm 0.02$} \\ 
  \hline \\ 
  All & $ 0.72$ \textcolor{gray}{$\pm 0.01$} & $ 0.73$ \textcolor{gray}{$\pm 0.01$} & $ 0.73$ \textcolor{gray}{$\pm 0.01$} & $ 0.65$ \textcolor{gray}{$\pm 0.02$} & $ 0.62$ \textcolor{gray}{$\pm 0.03$} \\ 
  \hline 
\end{tabular}
   \end{table}

 \begin{table}
   \caption{LSTM 200h results on Marvin \& Linzen subset without OOV words.}
   \label{tab:marvin-linzen-oov}
 \begin{tabular}{lrrrrr} 
  \hline \\ 
 & \multicolumn{1}{l}{nll} & \multicolumn{1}{l}{nll-neg-1.0} & \multicolumn{1}{l}{nll-neg-2.0} & \multicolumn{1}{l}{nll-neg-4.0} & \multicolumn{1}{l}{nll-neg-8.0} \\ 
  \hline \\ 
\textsc{Subject-verb agreement} & \multicolumn{4}{c}{} \\ 
  Simple & $ 1.00$ \textcolor{gray}{$\pm 0.00$} & $ 1.00$ \textcolor{gray}{$\pm 0.00$} & $ 1.00$ \textcolor{gray}{$\pm 0.00$} & $ 0.76$ \textcolor{gray}{$\pm 0.00$} & $ 0.81$ \textcolor{gray}{$\pm 0.00$} \\ 
  In a sentential complement & $ 0.89$ \textcolor{gray}{$\pm 0.03$} & $ 0.81$ \textcolor{gray}{$\pm 0.02$} & $ 0.89$ \textcolor{gray}{$\pm 0.05$} & $ 0.71$ \textcolor{gray}{$\pm 0.00$} & $ 0.69$ \textcolor{gray}{$\pm 0.00$} \\ 
  Short VP coordination & $ 0.99$ \textcolor{gray}{$\pm 0.01$} & $ 0.94$ \textcolor{gray}{$\pm 0.02$} & $ 0.99$ \textcolor{gray}{$\pm 0.01$} & $ 0.57$ \textcolor{gray}{$\pm 0.11$} & $ 0.69$ \textcolor{gray}{$\pm 0.00$} \\ 
  Long VP coordination & $ 0.65$ \textcolor{gray}{$\pm 0.02$} & $ 0.59$ \textcolor{gray}{$\pm 0.02$} & $ 0.60$ \textcolor{gray}{$\pm 0.00$} & $ 0.51$ \textcolor{gray}{$\pm 0.03$} & $ 0.51$ \textcolor{gray}{$\pm 0.00$} \\ 
  \rowcolor{lightgray} Across a prepositional phrase & $ 0.96$ \textcolor{gray}{$\pm 0.01$} & $ 0.97$ \textcolor{gray}{$\pm 0.00$} & $ 0.97$ \textcolor{gray}{$\pm 0.01$} & $ 0.58$ \textcolor{gray}{$\pm 0.03$} & $ 0.61$ \textcolor{gray}{$\pm 0.00$} \\ 
  \rowcolor{lightgray} Across a subject relative clause & $ 0.95$ \textcolor{gray}{$\pm 0.01$} & $ 0.98$ \textcolor{gray}{$\pm 0.01$} & $ 0.98$ \textcolor{gray}{$\pm 0.01$} & $ 0.62$ \textcolor{gray}{$\pm 0.05$} & $ 0.68$ \textcolor{gray}{$\pm 0.00$} \\ 
  \rowcolor{lightgray} Across an object relative clause & $ 0.72$ \textcolor{gray}{$\pm 0.03$} & $ 0.71$ \textcolor{gray}{$\pm 0.03$} & $ 0.79$ \textcolor{gray}{$\pm 0.06$} & $ 0.52$ \textcolor{gray}{$\pm 0.01$} & $ 0.57$ \textcolor{gray}{$\pm 0.00$} \\ 
  \rowcolor{lightgray} Across an objective relative (no that) & $ 0.67$ \textcolor{gray}{$\pm 0.02$} & $ 0.73$ \textcolor{gray}{$\pm 0.02$} & $ 0.74$ \textcolor{gray}{$\pm 0.01$} & $ 0.46$ \textcolor{gray}{$\pm 0.02$} & $ 0.54$ \textcolor{gray}{$\pm 0.00$} \\ 
  In an object relative clause & $ 0.64$ \textcolor{gray}{$\pm 0.01$} & $ 0.63$ \textcolor{gray}{$\pm 0.02$} & $ 0.64$ \textcolor{gray}{$\pm 0.01$} & $ 0.59$ \textcolor{gray}{$\pm 0.01$} & $ 0.59$ \textcolor{gray}{$\pm 0.00$} \\ 
  In an object relative (no that) & $ 0.65$ \textcolor{gray}{$\pm 0.01$} & $ 0.61$ \textcolor{gray}{$\pm 0.02$} & $ 0.60$ \textcolor{gray}{$\pm 0.04$} & $ 0.57$ \textcolor{gray}{$\pm 0.02$} & $ 0.63$ \textcolor{gray}{$\pm 0.00$} \\ 
  \hline \\ 
\textsc{Reflexive anaphora} & \multicolumn{4}{c}{} \\ 
  \rowcolor{lightgray} Simple (RA) & $ 0.84$ \textcolor{gray}{$\pm 0.03$} & $ 0.74$ \textcolor{gray}{$\pm 0.05$} & $ 0.87$ \textcolor{gray}{$\pm 0.04$} & $ 0.69$ \textcolor{gray}{$\pm 0.10$} & $ 0.57$ \textcolor{gray}{$\pm 0.00$} \\ 
  In a sentential complement & $ 0.89$ \textcolor{gray}{$\pm 0.03$} & $ 0.81$ \textcolor{gray}{$\pm 0.02$} & $ 0.89$ \textcolor{gray}{$\pm 0.05$} & $ 0.71$ \textcolor{gray}{$\pm 0.00$} & $ 0.69$ \textcolor{gray}{$\pm 0.00$} \\ 
  \rowcolor{lightgray} Across a relative clause (RA) & $ 0.72$ \textcolor{gray}{$\pm 0.02$} & $ 0.72$ \textcolor{gray}{$\pm 0.04$} & $ 0.77$ \textcolor{gray}{$\pm 0.01$} & $ 0.70$ \textcolor{gray}{$\pm 0.00$} & $ 0.71$ \textcolor{gray}{$\pm 0.00$} \\ 
  \hline \\ 
\textsc{Negative polarity items} & \multicolumn{4}{c}{} \\ 
  \rowcolor{lightgray} Simple (NPI) & $ 0.54$ \textcolor{gray}{$\pm 0.01$} & $ 0.54$ \textcolor{gray}{$\pm 0.02$} & $ 0.61$ \textcolor{gray}{$\pm 0.04$} & $ 0.55$ \textcolor{gray}{$\pm 0.09$} & $ 0.51$ \textcolor{gray}{$\pm 0.00$} \\ 
  \rowcolor{lightgray} Across a relative clause (NPI) & $ 0.56$ \textcolor{gray}{$\pm 0.01$} & $ 0.59$ \textcolor{gray}{$\pm 0.01$} & $ 0.60$ \textcolor{gray}{$\pm 0.01$} & $ 0.57$ \textcolor{gray}{$\pm 0.02$} & $ 0.55$ \textcolor{gray}{$\pm 0.00$} \\ 
  \hline \\ 
  \rowcolor{lightgray} All & $ 0.72$ \textcolor{gray}{$\pm 0.01$} & $ 0.72$ \textcolor{gray}{$\pm 0.00$} & $ 0.74$ \textcolor{gray}{$\pm 0.00$} & $ 0.56$ \textcolor{gray}{$\pm 0.01$} & $ 0.60$ \textcolor{gray}{$\pm 0.00$} \\ 
  \hline 
\end{tabular}
 \end{table}

 \begin{table}
   \caption{LSTM 200h results on all Marvin \& Linzen data \cite{marvin2018targeted}.}
   \label{tab:marvin-linzen-all}
 \begin{tabular}{lrrrrr} 
  \hline \\ 
 & \multicolumn{1}{l}{nll} & \multicolumn{1}{l}{nll-neg-1.0} & \multicolumn{1}{l}{nll-neg-2.0} & \multicolumn{1}{l}{nll-neg-4.0} & \multicolumn{1}{l}{nll-neg-8.0} \\ 
  \hline \\ 
\textsc{Subject-verb agreement} & \multicolumn{4}{c}{} \\ 
  Simple & $ 0.74$ \textcolor{gray}{$\pm 0.01$} & $ 0.73$ \textcolor{gray}{$\pm 0.01$} & $ 0.75$ \textcolor{gray}{$\pm 0.01$} & $ 0.73$ \textcolor{gray}{$\pm 0.01$} & $ 0.73$ \textcolor{gray}{$\pm 0.01$} \\ 
  In a sentential complement & $ 0.75$ \textcolor{gray}{$\pm 0.02$} & $ 0.71$ \textcolor{gray}{$\pm 0.02$} & $ 0.73$ \textcolor{gray}{$\pm 0.02$} & $ 0.72$ \textcolor{gray}{$\pm 0.01$} & $ 0.69$ \textcolor{gray}{$\pm 0.02$} \\ 
  \rowcolor{lightgray} Short VP coordination & $ 0.72$ \textcolor{gray}{$\pm 0.01$} & $ 0.71$ \textcolor{gray}{$\pm 0.01$} & $ 0.72$ \textcolor{gray}{$\pm 0.01$} & $ 0.74$ \textcolor{gray}{$\pm 0.01$} & $ 0.73$ \textcolor{gray}{$\pm 0.01$} \\ 
  Long VP coordination & $ 0.63$ \textcolor{gray}{$\pm 0.02$} & $ 0.61$ \textcolor{gray}{$\pm 0.01$} & $ 0.62$ \textcolor{gray}{$\pm 0.02$} & $ 0.60$ \textcolor{gray}{$\pm 0.01$} & $ 0.58$ \textcolor{gray}{$\pm 0.01$} \\ 
  \rowcolor{lightgray} Across a prepositional phrase & $ 0.76$ \textcolor{gray}{$\pm 0.00$} & $ 0.77$ \textcolor{gray}{$\pm 0.00$} & $ 0.78$ \textcolor{gray}{$\pm 0.00$} & $ 0.78$ \textcolor{gray}{$\pm 0.00$} & $ 0.78$ \textcolor{gray}{$\pm 0.00$} \\ 
  \rowcolor{lightgray} Across a subject relative clause & $ 0.69$ \textcolor{gray}{$\pm 0.01$} & $ 0.72$ \textcolor{gray}{$\pm 0.01$} & $ 0.73$ \textcolor{gray}{$\pm 0.00$} & $ 0.73$ \textcolor{gray}{$\pm 0.00$} & $ 0.72$ \textcolor{gray}{$\pm 0.00$} \\ 
  \rowcolor{lightgray} Across an object relative clause & $ 0.65$ \textcolor{gray}{$\pm 0.02$} & $ 0.71$ \textcolor{gray}{$\pm 0.02$} & $ 0.69$ \textcolor{gray}{$\pm 0.02$} & $ 0.72$ \textcolor{gray}{$\pm 0.02$} & $ 0.76$ \textcolor{gray}{$\pm 0.01$} \\ 
  \rowcolor{lightgray} Across an objective relative (no that) & $ 0.63$ \textcolor{gray}{$\pm 0.02$} & $ 0.71$ \textcolor{gray}{$\pm 0.01$} & $ 0.71$ \textcolor{gray}{$\pm 0.02$} & $ 0.72$ \textcolor{gray}{$\pm 0.01$} & $ 0.74$ \textcolor{gray}{$\pm 0.02$} \\ 
  In an object relative clause & $ 0.60$ \textcolor{gray}{$\pm 0.01$} & $ 0.60$ \textcolor{gray}{$\pm 0.01$} & $ 0.59$ \textcolor{gray}{$\pm 0.00$} & $ 0.57$ \textcolor{gray}{$\pm 0.01$} & $ 0.55$ \textcolor{gray}{$\pm 0.01$} \\ 
  In an object relative (no that) & $ 0.60$ \textcolor{gray}{$\pm 0.01$} & $ 0.60$ \textcolor{gray}{$\pm 0.01$} & $ 0.60$ \textcolor{gray}{$\pm 0.01$} & $ 0.58$ \textcolor{gray}{$\pm 0.01$} & $ 0.57$ \textcolor{gray}{$\pm 0.01$} \\ 
  \hline \\ 
\textsc{Reflexive anaphora} & \multicolumn{4}{c}{} \\ 
  Simple (RA) & $ 0.79$ \textcolor{gray}{$\pm 0.02$} & $ 0.70$ \textcolor{gray}{$\pm 0.03$} & $ 0.72$ \textcolor{gray}{$\pm 0.03$} & $ 0.67$ \textcolor{gray}{$\pm 0.03$} & $ 0.63$ \textcolor{gray}{$\pm 0.03$} \\ 
  In a sentential complement & $ 0.75$ \textcolor{gray}{$\pm 0.02$} & $ 0.71$ \textcolor{gray}{$\pm 0.02$} & $ 0.73$ \textcolor{gray}{$\pm 0.02$} & $ 0.72$ \textcolor{gray}{$\pm 0.01$} & $ 0.69$ \textcolor{gray}{$\pm 0.02$} \\ 
  Across a relative clause (RA) & $ 0.69$ \textcolor{gray}{$\pm 0.02$} & $ 0.67$ \textcolor{gray}{$\pm 0.01$} & $ 0.65$ \textcolor{gray}{$\pm 0.01$} & $ 0.67$ \textcolor{gray}{$\pm 0.01$} & $ 0.69$ \textcolor{gray}{$\pm 0.03$} \\ 
  \hline \\ 
\textsc{Negative polarity items} & \multicolumn{4}{c}{} \\ 
  Simple (NPI) & $ 0.57$ \textcolor{gray}{$\pm 0.01$} & $ 0.57$ \textcolor{gray}{$\pm 0.01$} & $ 0.56$ \textcolor{gray}{$\pm 0.01$} & $ 0.58$ \textcolor{gray}{$\pm 0.01$} & $ 0.55$ \textcolor{gray}{$\pm 0.01$} \\ 
  Across a relative clause (NPI) & $ 0.58$ \textcolor{gray}{$\pm 0.00$} & $ 0.58$ \textcolor{gray}{$\pm 0.01$} & $ 0.58$ \textcolor{gray}{$\pm 0.00$} & $ 0.57$ \textcolor{gray}{$\pm 0.01$} & $ 0.57$ \textcolor{gray}{$\pm 0.01$} \\ 
  \hline \\ 
  \rowcolor{lightgray} All & $ 0.65$ \textcolor{gray}{$\pm 0.01$} & $ 0.67$ \textcolor{gray}{$\pm 0.00$} & $ 0.67$ \textcolor{gray}{$\pm 0.00$} & $ 0.67$ \textcolor{gray}{$\pm 0.00$} & $ 0.67$ \textcolor{gray}{$\pm 0.01$} \\ 
  \hline 
\end{tabular}
 \end{table}

 \section{Negative data examples}
 \label{app:negative-data}

Here we provide some examples of the type of negative word-based tri-gram data we employ
for training and evaluation.  In
Figure~\ref{fig:negative-examples-dev-ptb} we show a few random
negative sentences from a model trained on the PTB dev set.  In
Figure~\ref{fig:negative-examples-dev-sva}, we show a few random
negative sentences from a mdoel trained on the subject-verb agreement
dev set.  Some sentences resemble language, while others are complete
gibberish.  The training sentences tend to be more language like since
those models are always pre-conditioned on appropriate context.  The
dev-data is generated in a single-go without re-conditioning.
 
\begin{figure}[h]
  \centering
  \begin{tabular}{|p{0.9\textwidth}|}
    \hline
 for instance for a piece on local tv stations and sells some of its \$ N 
 the dow jones transportation average second in size only to the customer .\\
 he visits the same time  .\\
 today of all the witnesses both congressmen and industry said  .\\
 the ads touted fidelity 's automated <unk> beneath the huge drop in stock prices firmed up again as traders sold big baskets of stock .\\
    moody 's investors service inc. downgraded its ratings on the chicago market makers to get advertisers to use their clout to help insure the integrity of the home fans in this country .\\
    \hline
    \end{tabular}
 \caption{Sampling of the tri-grams version of the official PTB dev set (with Mikolov's pre-processing).  The period indicates end of the line, N indicates a number and <unk> is a symbol that replaces the rare words in the original corpus, but that the tri-gram model still learns to generate.}
 \label{fig:negative-examples-dev-ptb}
 \end{figure}

\begin{figure}[h]
  \centering
  \begin{tabular}{|p{0.9\textwidth}|}
    \hline
NNP is the NN VBZ the post 's other major began `` VBG their way to prove that they face off against a sourced recreation with appropriate links to the NNS of flies or NNS on the west are available . \\
when a spell NN , brain waves and body movements while they may contain useful info , 5 , and as strong as physical force . \\
the arms show 17 gold NNS on the wars of that name is just east of the increased potential energy .
and if NNS , and when at the temple . \\
the saxon NNP river then flows from NNP . \\
the highest point in the plane p NN to the north and south branch , ford branch , both of these two lines . \\
on the western edge of NNP . \\
collaboration generally is a combination of natural circulation include NNS and of the VBG fan input or adjacent to the west represent NNP NNS , NNS and NNP ) uses two proprietary standards instead of four tunnels that have common elements from all major NNS ; examples include red NNP , NNP NNP , st . \\
only a few sandy beaches , the NN and a bit too much  \\
    \hline
    \end{tabular}
 \caption{Tri-gram version of Linzen\&Goldberg subject-verb agreement data set.  The tri-gram is trained on the development portion of the data.  Rare words are replaced with their part of speech: NN is a singular noun, NNS a plural noun, NNP a singular proper noun, VBZ a present singular verb, VBG a gerund, etc.}
 \label{fig:negative-examples-dev-sva}
 \end{figure}

\section{Fine-tuning by attenuating n-gram statistics}
\label{app:gpt2}
In the experiments above, we demonstrated that language model training by attenuating tri-gram statistics improves models' ability to both rule out "negative" sentences (Section~\ref{sec:detecting}) and navigate long-distance syntactic dependencies (Section~\ref{sec:sva}) across a variety of architectures, including GPT2. 

In this section, we explore fine-tuning GPT2 using our method in greater detail. Overall, our results indicate that tri-gram attenuation a) increases negative-data perplexity without harming positive-data perplexity (Section~\ref{sec:gpt2-detecting}) and b) improves GPT2's performance across most of the long-distance constructions in an additional syntactic evaluation task (Section~\ref{sec:gpt2-syntax}). This serves as preliminary evidence that n-gram attenuation during fine-tuning is an effective way to improve GPT2's interpretation of long-distance dependencies without sacrificing language model quality and without using additional data beyond the fine-tuning set itself. We leave assessment of other Transformer architectures, as well as a detailed examination of the interaction between n-gram order and model quality, for future work.

\subsection{Model \& fine-tuning details}
\label{app:gpt2-details}
We fine-tune the base GPT2 model (110M parameters) via the HuggingFace library \cite{Wolf2019HuggingFacesTS} \footnote{Release 2.8.0}. In all cases, we fine-tune for 10 epochs using the Adam optimizer \cite{DBLP:journals/corr/KingmaB14} with decoupled weight decay regularization \cite{loshchilov2018decoupled} as implemented in \cite{Wolf2019HuggingFacesTS}. We use an initial learning rate of $5e-5$, $\epsilon = 1e-8$, and clip gradients at $1$. We did not tune these hyperparameters, but rather used the default values given in HuggingFace. 

For all datasets, we use the same train/dev/test splits as described above. For the Penn Treebank experiments in Section~\ref{sec:gpt2-detecting}, we use batches of size $16$ over text windows of length $20$ tokens. In all other experiments, we use batches of $32$ sentences. As in prior experiments, during training, we generate a "negative" version of each batch using an unsmoothed tri-gram language model trained beforehand using maximum likelihood estimation \cite{loper2002nltk}. 

For each experiment, we report the results for a single fine-tuning/evaluation run. We leave robust estimation of invariance to random seed, settings of $\alpha$, and other hyperparameters for future work.

\subsection{Results: detecting and attenuating n-grams}
\label{sec:gpt2-detecting}
\begin{figure}[h]
    \centering
    \begin{subfigure}[b]{0.32\textwidth}
        \centering
        \includegraphics[width=\textwidth]{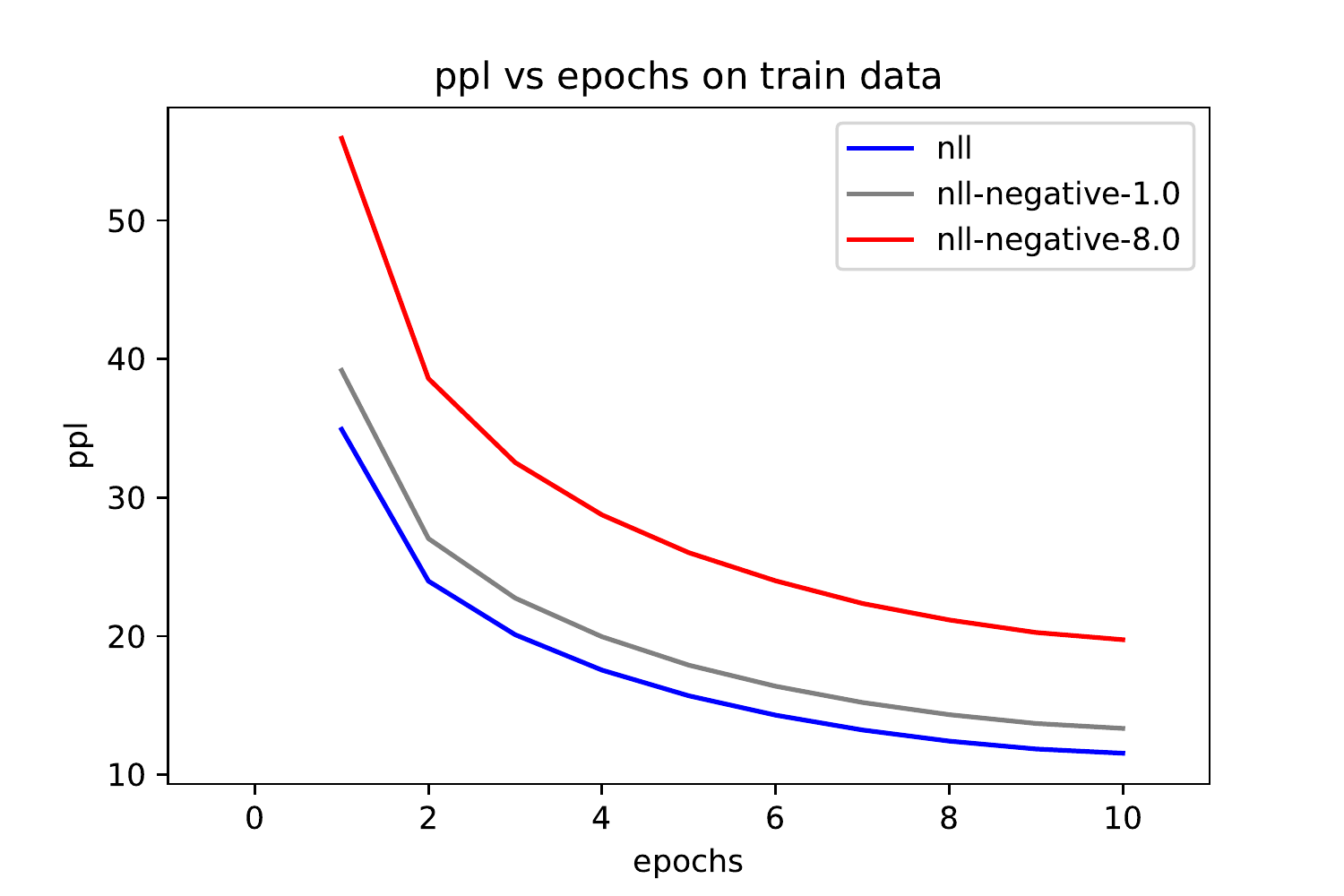}
        \caption{Train perplexity.}
        \label{fig:gpt2-train-ppl}
    \end{subfigure}
    \begin{subfigure}[b]{0.32\textwidth}
        \centering
        \includegraphics[width=\textwidth]{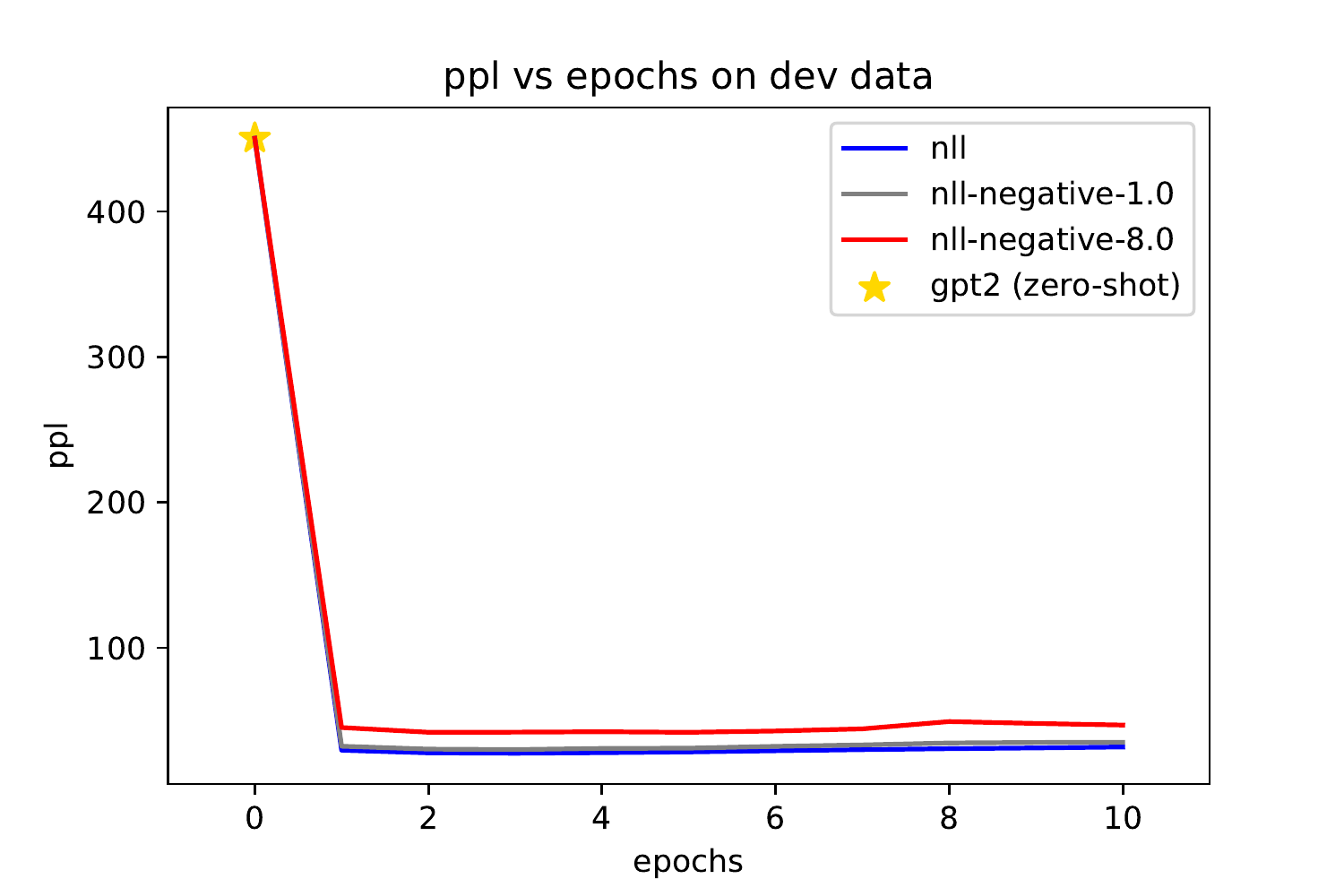}
        \caption{Dev perplexity (positive data).}
        \label{fig:gpt2-dev-ppl}
    \end{subfigure}
    \begin{subfigure}[b]{0.32\textwidth}
        \centering
        \includegraphics[width=\textwidth]{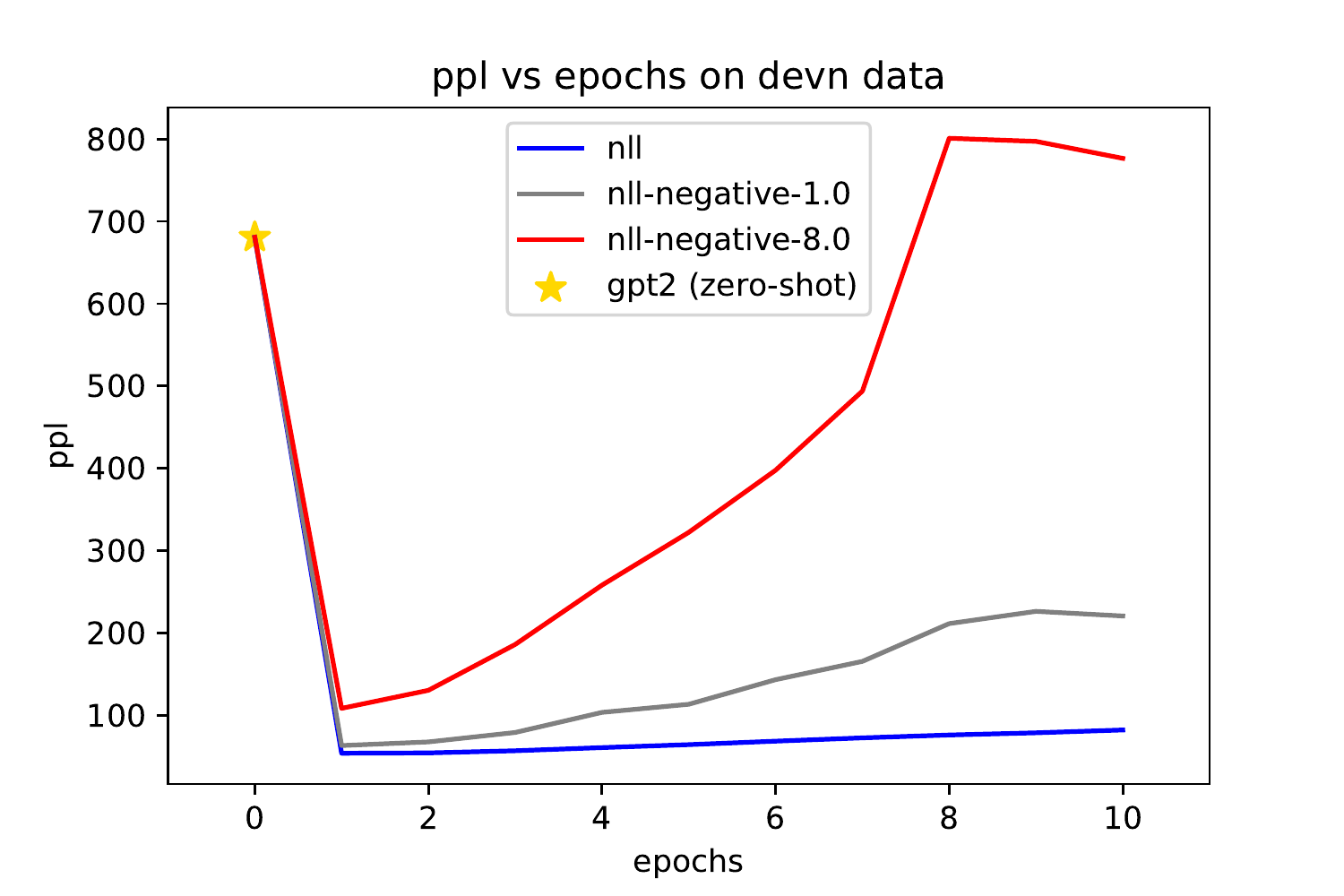}
        \caption{Dev perplexity (negative data).}
        \label{fig:gpt2-devn-ppl}
    \end{subfigure}
    \caption{Perplexity on train, dev, and negative dev data (GPT-2).
        \label{fig:gpt2-ptb-ppl}}
\end{figure}

To inspect how tri-gram attenuation affects language model quality as fine-tuning progresses, we measure perplexity on the PTB train, dev, and "negative dev" sets (described in Section~\ref{sec:data}) after each training epoch. As was the case with the LSTMs, "positive" dev perplexity stays constant while the the negative-data objective causes "negative" dev perplexity to increase in the cases where $\alpha > 0$ was used (Fig~\ref{fig:gpt2-ptb-ppl}).

\subsection{Results on targeted syntactic evaluation}
\label{sec:gpt2-syntax}
To evaluate whether fine-tuning with exorcism improves GPT2's syntactic abilities, we fine-tune on the subject-verb agreement training set provided by \cite{linzen16assessing} and, as in Section~\ref{app:data}, we evaluate on the syntactic constructions from \cite{marvin2018targeted}.

Table~\ref{tab:gpt2-marvin-linzen-all-fixed} shows that fine-tuning with tri-gram attenuation improves the language model's interpretation of many constructions. Most of the cases where our method hurt performance involve short-distance dependencies. For example, as noted in \cite{marvin2018targeted}, dependencies within object relative clauses are local:

\eenumsentence{
\item  \label{ex:farmera} The farmer that the parents \underline{love} swims.
\item \label{ex:farmerb} *The farmer that the parents \underline{loves} swims.
}

That is, "parents" and "love" must agree; the distractor is "farmer". Such cases show that exorcising away n-gram statistics can cause language models to misinterpret dependencies that are close together. Indeed, performance in both "within object RC" categories more-or-less decreased as exorcism strength ($\alpha$) increased. This suggests that n-gram attenuation during fine-tuning introduces a trade-off between optimizing the model to interpret long-distance dependencies at the expense of interpreting short-distance dependencies. We leave further investigation of this trade-off for future work.

\begin{table}[h]
    \caption{GPT2 results on Marvin \& Linzen data \cite{marvin2018targeted}. Highlighted rows indicate those for which the mean of the "negative data" results mean(nll-neg-1, nll-neg-8) beat the maximum baseline score max(gpt2, nll).}
    \label{tab:gpt2-marvin-linzen-all-fixed}
\begin{tabular}{lrrrr}
  \hline \\
 & \multicolumn{1}{l}{gpt2 (zero-shot)} & \multicolumn{1}{l}{nll} & \multicolumn{1}{l}{nll-neg-1} & \multicolumn{1}{l}{nll-neg-8} \\
\hline \\
\textsc{Subject-verb agreement} & \multicolumn{3}{c}{} \\
Simple & $ 0.89$ \textcolor{gray}{$\pm 0.00$} & $ \mathbf{0.99}$ \textcolor{gray}{$\pm 0.00$} & $ \mathbf{0.99}$ \textcolor{gray}{$\pm 0.00$} & $ 0.94$ \textcolor{gray}{$\pm 0.00$} \\
In a sentential complement & $ 0.84$ \textcolor{gray}{$\pm 0.00$} & $ 0.92$ \textcolor{gray}{$\pm 0.00$} & $ \mathbf{0.93}$ \textcolor{gray}{$\pm 0.00$} & $ 0.91$ \textcolor{gray}{$\pm 0.00$} \\
Short VP coordination & $ \mathbf{0.96}$ \textcolor{gray}{$\pm 0.00$} & $ \mathbf{0.96}$ \textcolor{gray}{$\pm 0.00$} & $ 0.95$ \textcolor{gray}{$\pm 0.00$} & $ 0.94$ \textcolor{gray}{$\pm 0.00$} \\
\rowcolor{lightgray} Long VP coordination & $ 0.91$ \textcolor{gray}{$\pm 0.00$} & $ 0.93$ \textcolor{gray}{$\pm 0.00$} & $ 0.93$ \textcolor{gray}{$\pm 0.00$} & $ \mathbf{0.94}$ \textcolor{gray}{$\pm 0.00$} \\
Across a prepositional phrase & $ 0.77$ \textcolor{gray}{$\pm 0.00$} & $ \mathbf{0.95}$ \textcolor{gray}{$\pm 0.00$} & $ \mathbf{0.95}$ \textcolor{gray}{$\pm 0.00$} & $ \mathbf{0.95}$ \textcolor{gray}{$\pm 0.00$} \\
Across a subject relative clause & $ 0.72$ \textcolor{gray}{$\pm 0.00$} & $ 0.93$ \textcolor{gray}{$\pm 0.00$} & $ 0.92$ \textcolor{gray}{$\pm 0.00$} & $ \mathbf{0.94}$ \textcolor{gray}{$\pm 0.00$} \\
\rowcolor{lightgray} Across an object relative clause & $ 0.84$ \textcolor{gray}{$\pm 0.00$} & $ 0.92$ \textcolor{gray}{$\pm 0.00$} & $ 0.93$ \textcolor{gray}{$\pm 0.00$} & $ \mathbf{0.94}$ \textcolor{gray}{$\pm 0.00$} \\
\rowcolor{lightgray} Across an objective relative (no that) & $ 0.77$ \textcolor{gray}{$\pm 0.00$} & $ 0.88$ \textcolor{gray}{$\pm 0.00$} & $ \mathbf{0.89}$ \textcolor{gray}{$\pm 0.00$} & $ \mathbf{0.89}$ \textcolor{gray}{$\pm 0.00$} \\
In an object relative clause & $ \mathbf{0.94}$ \textcolor{gray}{$\pm 0.00$} & $ 0.79$ \textcolor{gray}{$\pm 0.00$} & $ 0.75$ \textcolor{gray}{$\pm 0.00$} & $ 0.77$ \textcolor{gray}{$\pm 0.00$} \\
In an object relative (no that) & $ \mathbf{0.89}$ \textcolor{gray}{$\pm 0.00$} & $ 0.77$ \textcolor{gray}{$\pm 0.00$} & $ 0.75$ \textcolor{gray}{$\pm 0.00$} & $ 0.75$ \textcolor{gray}{$\pm 0.00$} \\
\hline \\
\textsc{Reflexive anaphora} & \multicolumn{3}{c}{} \\
Simple (RA) & $ \mathbf{0.98}$ \textcolor{gray}{$\pm 0.00$} & $ 0.92$ \textcolor{gray}{$\pm 0.00$} & $ 0.94$ \textcolor{gray}{$\pm 0.00$} & $ 0.94$ \textcolor{gray}{$\pm 0.00$} \\
In a sentential complement (RA) & $ \mathbf{0.88}$ \textcolor{gray}{$\pm 0.00$} & $ 0.83$ \textcolor{gray}{$\pm 0.00$} & $ 0.81$ \textcolor{gray}{$\pm 0.00$} & $ 0.82$ \textcolor{gray}{$\pm 0.00$} \\
\rowcolor{lightgray} Across a relative clause (RA) & $ 0.76$ \textcolor{gray}{$\pm 0.00$} & $ 0.76$ \textcolor{gray}{$\pm 0.00$} & $ 0.75$ \textcolor{gray}{$\pm 0.00$} & $ \mathbf{0.79}$ \textcolor{gray}{$\pm 0.00$} \\
\hline \\
\textsc{Negative polarity items} & \multicolumn{3}{c}{} \\
\rowcolor{lightgray} Simple (NPI) & $ 0.93$ \textcolor{gray}{$\pm 0.00$} & $ 0.96$ \textcolor{gray}{$\pm 0.00$} & $ \mathbf{0.98}$ \textcolor{gray}{$\pm 0.00$} & $ \mathbf{0.98}$ \textcolor{gray}{$\pm 0.00$} \\
\rowcolor{lightgray} Across a relative clause (NPI) & $ 0.83$ \textcolor{gray}{$\pm 0.00$} & $ 0.91$ \textcolor{gray}{$\pm 0.00$} & $ 0.90$ \textcolor{gray}{$\pm 0.00$} & $ \mathbf{0.94}$ \textcolor{gray}{$\pm 0.00$} \\
\hline \\
All & $ 0.82$ \textcolor{gray}{$\pm 0.00$} & $ 0.86$ \textcolor{gray}{$\pm 0.00$} & $ 0.85$ \textcolor{gray}{$\pm 0.00$} & $ \mathbf{0.87}$ \textcolor{gray}{$\pm 0.00$} \\
  \hline
\end{tabular}
\end{table}

\end{document}